\documentclass[10pt,twocolumn,letterpaper]{article}

\usepackage[pagenumbers]{cvpr}      
\usepackage{float}
\usepackage{times}
\usepackage{epsfig}
\usepackage{caption}
\usepackage{graphicx}
\usepackage{amsmath}
\usepackage{amssymb}
\usepackage{bbm}
\usepackage{multirow}
\usepackage{booktabs}
\usepackage{lipsum}  
\usepackage{wrapfig}
\usepackage{xcolor}
\usepackage{wrapfig}
\usepackage[dvipsnames]{xcolor}

\newcommand\blfootnote[1]{%
  \begingroup
  \renewcommand\thefootnote{}\footnote{#1}%
  \addtocounter{footnote}{-1}%
  \endgroup
}
\definecolor{cvprblue}{rgb}{0.21,0.49,0.74}
\usepackage[pagebackref,breaklinks,colorlinks,allcolors=cvprblue]{hyperref}

\def\paperID{} 
\def\confName{CVPR}
\def\confYear{2026}

\title{Image2Garment: Simulation-ready Garment Generation from a Single Image}

\author{Selim Emir Can$^{1*}$ $\quad\quad\quad$
Jan Ackermann$^{1*}$ $\quad\quad\quad$
Kiyohiro Nakayama$^{1}$\\
Ruofan Liu$^{3,1}$ $\quad\quad\quad$
Tong Wu$^1$ $\quad\quad\quad$
Yang Zheng$^1$ $\quad\quad\quad$
Hugo Bertiche$^2$\\
Menglei Chai$^2$ $\quad\quad\quad$
Thabo Beeler$^2$ $\quad\quad\quad$
Gordon Wetzstein$^1$\\[0.3cm]
$\phantom{}^1$Stanford University $\quad\quad\quad$ $\phantom{}^2$Google $\quad\quad\quad$
$\phantom{}^3$Institute of Science Tokyo
}

\begin{document}

\twocolumn[{%
  \begin{center}
    \renewcommand\twocolumn[1][]{#1}%
    \maketitle
    \centering
    \includegraphics[width=\textwidth]{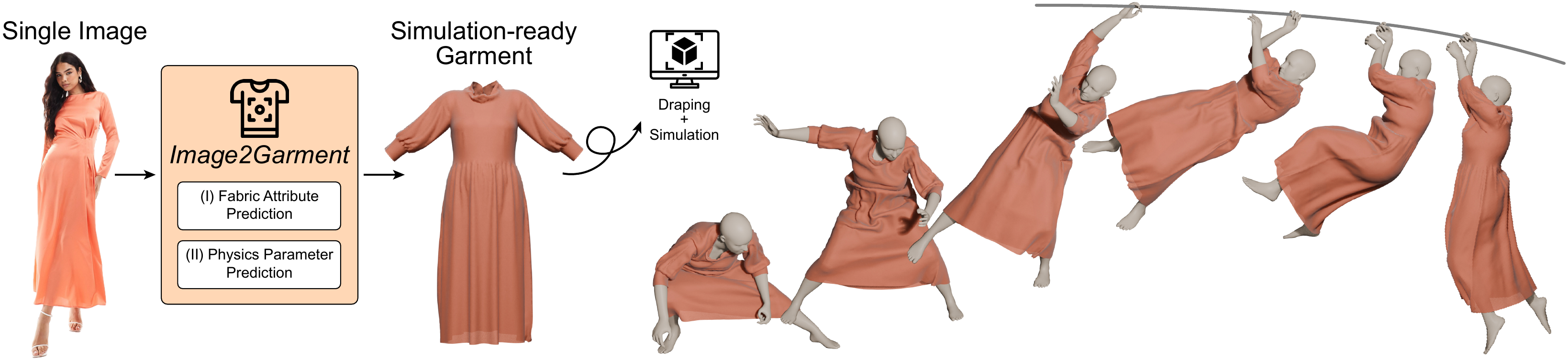}
    \captionof{figure}{Image2Garment is a feedforward framework that generates simulation-ready garments from a single image. For this purpose, garment geometry and physical fabric parameters are jointly predicted and used to simulate garment dynamics.}
    \label{fig:teaser}
  \end{center}
}]

\blfootnote{${}^*$ denotes equal contribution.}

\begin{abstract}
Estimating physically accurate, simulation-ready garments from a single image is challenging due to the absence of image-to-physics datasets and the ill-posed nature of this problem. Prior methods either require multi-view capture and expensive differentiable simulation or predict only garment geometry without the material properties required for realistic simulation. We propose a feed-forward framework that sidesteps these limitations by first fine-tuning a vision–language model to infer material composition and fabric attributes from real images, and then training a light-weight predictor that maps these attributes to the corresponding physical fabric parameters using a small dataset of material–physics measurements.
Our approach introduces two new datasets (FTAG and T2P) and delivers simulation-ready garments from a single image without iterative optimization. Experiments show that our estimator achieves superior accuracy in material composition estimation and fabric attribute prediction, and by passing them through our physics parameter estimator, we further achieve higher fidelity simulations compared to state-of-the-art image-to-garment methods. The project website is at \url{https://image2garment.github.io/}.
\end{abstract}
\vspace{-12px}   
\section{Introduction}
\label{sec:intro}
Generating simulation-ready garments from visual observations is increasingly important for applications in virtual reality, gaming, and fashion design. Although recent methods can recover physical fabric properties, they typically rely on multi-view capture setups or manual material measurement, both of which are labor-intensive and impractical outside controlled environments. While single-image inference would be an accessible alternative, estimating a garment’s physical properties (e.g., stretch and bend stiffness, density, damping) from an in-the-wild image remains profoundly challenging due to limited viewpoints, ambiguous drape cues, and the lack of direct supervision.

\begin{figure}[t]
    \centering
    \begin{tabular}{@{}cccc@{}}
        \includegraphics[width=.25\linewidth]{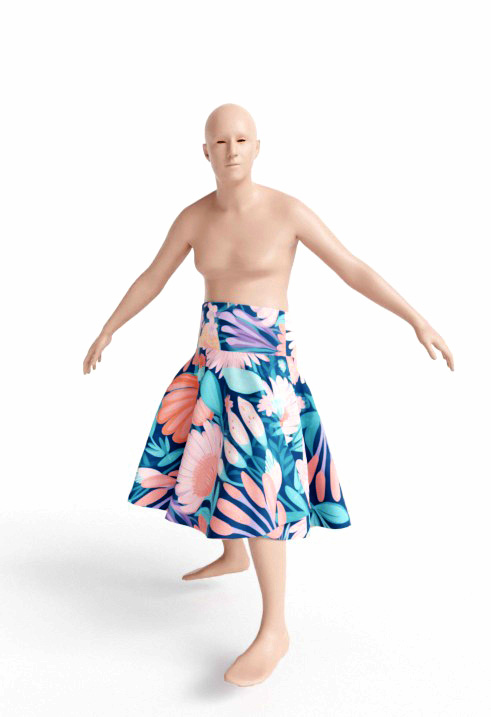}%
        \includegraphics[width=.25\linewidth]{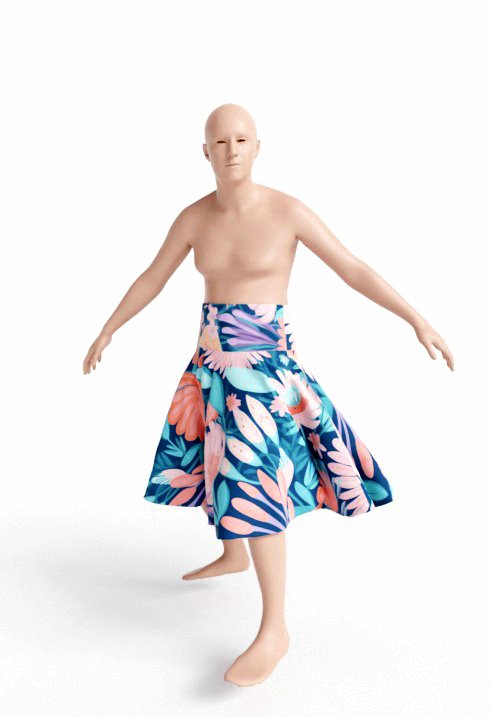}%
        \includegraphics[width=.25\linewidth]{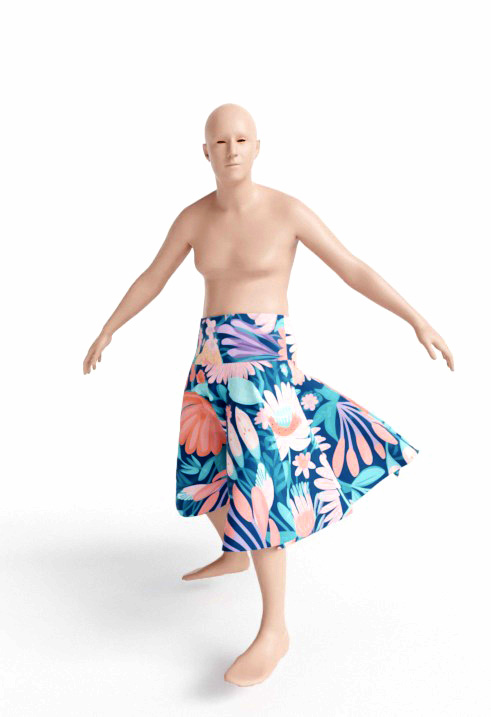}%
        \includegraphics[width=.25\linewidth]{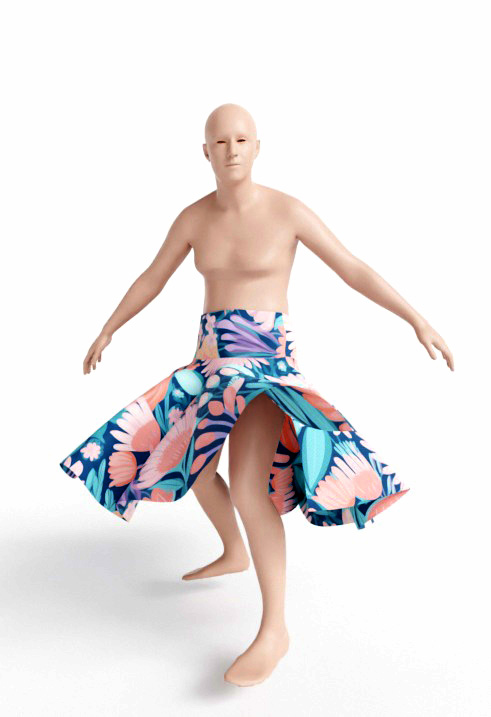} \\
        \multicolumn{4}{c}{%
    \hspace{-0.8em}\scriptsize\begin{tabular}{@{}c@{\hspace{0.5em}}c@{\hspace{0.5em}}c@{\hspace{0.5em}}c@{}}
        100\% Wool, Knit, & 70\% Cork 30\% Cotton, & 100\% Polyester, & Random \\
        Fleece, 382.32 gsm, & Knit, Lace, 16.00 gsm, & Woven, Twill, & Physics Params. \\
        1.05 mm & 0.5 mm & 195.00 gsm, 0.65 mm & 
    \end{tabular}
}
    \end{tabular}
    \caption{\textbf{Impact of garment fabric parameters on simulation.} We visualize the final frame of a jumping animation for four different fabrics (wool, cork--cotton, polyester, and a random material) each starting from the exact same initial condition. The choice of garment physics parameters changes the dynamics of the animation drastically. In turn, this makes it critical to estimate these parameters accurately in image-to-garment generation settings to faithfully predict shape, appearance, and dynamics of a garment from visual observations.}
    \vspace{-18pt}
    \label{fig:physics_parameter_importance}
\end{figure}

Recent progress in single-image garment generation, driven by fine-tuning Vision-Language Models (VLMs), has enabled substantial improvements in predicting garment geometry and appearance~\cite{bian2025chatgarmentgarmentestimationgeneration,nakayama2025aipparel,zhou2025design2garmentcode,liu2024multimodallatentdiffusionmodel,li2025dress}. However, these methods largely neglect physical parameters necessary for faithful simulation~(see Fig.~\ref{fig:physics_parameter_importance}).
Conversely, methods that optimize physical parameters via a differentiable simulation~\cite{zheng2024physavatar,rong2024gaussiangarments,li2024diffavatar,guo2025pgc,li2025dress} require multi-view inputs, involve slow iterative optimization, and are typically restricted to simple garment categories, making them impractical for real-world deployment. As a result, obtaining simulation-ready garments from a single image remains a challenging problem. While optimization-based methods struggle with the limited supervision available from a single image, directly training a neural network for this task is equally infeasible due to the lack of appropriate data. No large-scale dataset that pairs real-world garment images with the physical parameters required by cloth simulators exists, and creating such a resource would demand prohibitive manual effort.

Our key insight is to reformulate this inverse problem through a semantically grounded latent decomposition. Although direct image-to-physics supervision is unavailable, image-to-material information is abundant. Large online catalogs provide reliable material-composition labels (e.g., cotton, silk, polyester blends) as well as complementary garment attributes such as fabric family, weave structure, and thickness indicators. Importantly, these descriptors occupy a structured and relatively low-dimensional space that has a far more predictable relationship to the physical parameters used in cloth simulation. As a result, the material-to-physics mapping is substantially easier to learn and requires only a modest amount of material–physics data that can realistically be collected from industry-standard simulators.

Building on this observation, we introduce a two-stage factorization. First, we train a model to infer an interpretable set of material descriptors from the single input image. Second, using a compact set of independently gathered material–physics annotations, we train a mapper that converts these descriptors into the full set of simulator parameters. This latent-variable formulation regularizes the learning task, dramatically reduces data requirements, and resolves the ill-posedness that makes direct image-to-physics prediction impractical. The result is a fully feed-forward, optimization-free image-to-simulation pipeline.

In summary our contributions are:
\begin{itemize}
    \item Two datasets: (1) The Fabric Attributes from Garment Tags (FTAG) dataset—containing images annotated with material composition, fabric family, and structure type---and (2) the Tag-to-Physics (T2P) Dataset which links fabric attributes with measurable physical parameters directly compatible with an industry standard simulator.
    \item A feed-forward framework for single-image to simulation-ready garment generation, jointly predicting garment geometry and interpretable fabric descriptors, and mapping them to physically realistic material parameters.
    \item Extensive experiments demonstrating superior accuracy and speed, showing that our approach predicts accurate garment physics parameters that can be directly plugged in to image-to-simulation workflows.
\end{itemize}

\section{Related Work}
\label{sec:related}
\begin{figure*}[thb]
    \centering
    \includegraphics[width=\textwidth]{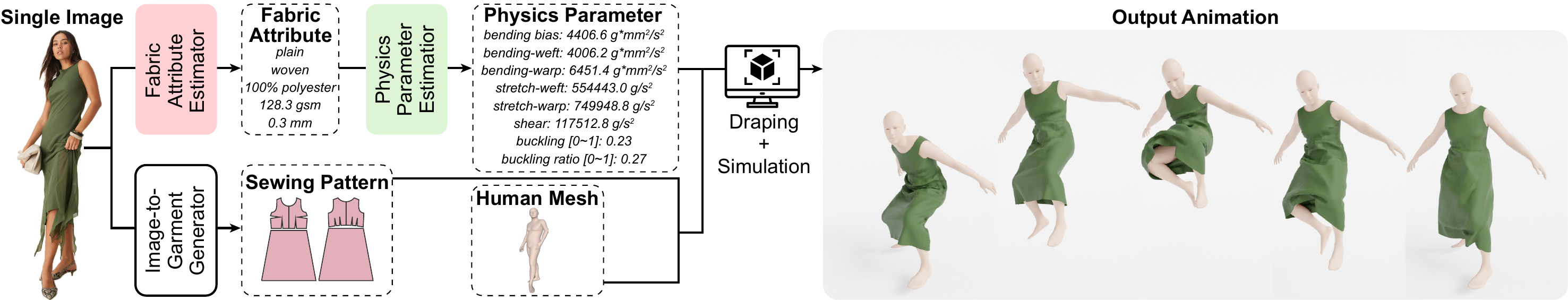}
    \caption{\textbf{Overview of the Image2Garment pipeline.} From a single image, we first generate the garment sewing pattern using ChatGarment~\cite{bian2025chatgarmentgarmentestimationgeneration}. Then we predict fabric attributes such as material composition, fabric family, structure type, weight and thickness aligned with standardized commercial garment tags. Finally, fabric physics parameters are estimated from the predicted attributes, following~\cite{dominguezelvira2024MechFromMet}, yielding mechanically interpretable quantities that describe fabric deformation. The garment geometry and physical parameters are then used to produce simulation-ready garments and physically accurate draping animations for any given motion sequence or body poses such as SMPL~\cite{SMPL:2015}.}
    \label{fig:pipeline}
\end{figure*}
\paragraph{Garment reconstruction and generation.}
Many works have explored how to reconstruct garments from multi-view images and videos. Works such as~\cite{Feng2022scarf, 11125958, jiang2022selfrecon, alldieck19cvpr, 8578973, bhatnagar2019mgn, NeuralTailor2022} take videos, multi-view images, or point clouds as input to recover the garment geometry. However, they do not model the material composition of the garments, and therefore cannot be directly used for physics simulation. Using inverse rendering and inverse physics, other approaches \cite{zheng2024physavatar, rong2024gaussiangarments, guo2025pgc, lee2025mpmavatarlearning3dgaussian, 11180929} optimize both the garment geometry and its physical parameters. However, they require studio-quality videos as input, making the algorithm impractical for casual users. 

More recently, generative models have enabled single-image-to-garment generation. Works such as~\cite{xiu2022icon, xiu2023econ, He_2021_ICCV, 9157750, alldieck2019tex2shape, 8885520, pifuSHNMKL19, 10.1007/978-3-030-58565-5_2} generate 3D meshes of garments fused with humans, requiring post-processing to separate the garment from the human mesh. Others~\cite{corona2021smplicit, Moon_2022_ECCV_ClothWild, ren2022dig, li2024garment, li2025single} model the human mesh separately from the garment, but still lack the necessary physics parameters for simulation. Another line of works~\cite{jeong_2015_garment_capture_from_photo, yang_2018_physics_inspired_garment_from_image, nakayama2025aipparel, bian2025chatgarmentgarmentestimationgeneration, liu2024multimodallatentdiffusionmodel, zhou2025design2garmentcode, liu2023sewformer, he2024dresscodeautoregressivelysewinggenerating, li2025garmagenet} directly generates the sewing pattern -- a CAD representation of garments -- from a single image by training on sewing pattern datasets~\cite{GarmentCodeData:2024, GarmentCode2023}. While this representation is more simulation ready as it automatically includes garment texture maps, users still need to specify physical parameters manually for each fabric panel. 

The work closest to ours is Dress 1-to-3~\cite{li2025dress}, which estimates both sewing pattern and physical parameter from a single image, leveraging inverse optimization. However, their method requires hours of optimization per garment and only outputs specific physical parameters that are not generalizable for other types of simulators. Moreover, code is not available for this method.

\paragraph{Optimizing physics parameters from images and videos.}
Estimating physics from visual observations has become an increasingly important topic in computer vision~\cite{battaglia2016interaction,chang2016compositional,watters2017visual,wu2017learning,wu2015galileo,lerer2016learning,groth2018shapestacks,yi2020clevrer,bakhtin2019phyre,garcia2025learning,kandukuri2022physical,ehrhardt2018unsupervised,le2025pixie,zhang2024physdreamer}.
Several works have also attempted to obtain physics parameters for garments from visual data~\cite{bhat2003cloth,feng2022learning,mao2020clothvideo,li2024diffavatar,yu2024inverse,santesteban2022snug,bertiche2022neuralcloth,li2024manifoldtransformer,ponsmoll2017clothcap,zheng2024physavatar,guo2025pgc,rong2024gaussiangarments,li2025dress}. However, all of these methods require strong visual supervision with multi-view or video setups and most are not compatible with established cloth simulators. This clearly shows the difficulty of predicting physics parameters usable in general physics simulators from a single image.

High-quality estimation of garment physical parameters has been studied primarily in controlled settings. Wang et al.~\cite{wang2011data} fit elastic models from measurements of real fabric samples, and Bhat et al.~\cite{bhat2003cloth} estimate parameters by matching simulated dynamics to video observations. The latter work refines these measurement-based approaches, including woven-fabric models~\cite{clyde2017woven}, friction estimation~\cite{rasheed2020learning}, and differentiable-physics pipelines using multi-view or depth data~\cite{gong2024bayesian,zheng2024differentiable,zheng2024physavatar}. These methods achieve high fidelity but depend on specialized capture setups. In contrast, we target parameter estimation from a single RGB image.

\section{Method}
\label{sec:datasets}

\paragraph{Overview.}
Our goal is to recover a \emph{simulation-ready garment} from a single RGB image \(I\).
We define a simulation-ready garment as \(G=(S,\boldsymbol{\theta})\), where \(S\) is the 3D garment shape and \(\boldsymbol{\theta}\) are the physical parameters required by a cloth simulator (e.g., stretch, bend, damping, friction).
We therefore seek to estimate
\begin{equation}
\hat{G} = \arg\max_{G} \, p(G \mid I) = \arg\max_{S,\boldsymbol{\theta}} \, p(S,\boldsymbol{\theta}\mid I).
\label{eq:goal}
\end{equation}

While single-view shape estimation \(p(S\mid I)\) has seen rapid progress, estimating \(\boldsymbol{\theta}\) directly from \(I\) is underexplored: paired image-to-physics data do not exist and the mapping is ambiguous. 
Our key insight is to introduce a latent variable, material \(M\) (and auxiliary garment attributes \(Z\); e.g., weave, thickness, finish), and decompose the posterior as\footnote{We make the assumption that the garment geometry $S$ and physical parameters $\boldsymbol{\theta}$ are conditionally independent given the input image $I$.}
\begin{equation}
\begin{split}
    p(S,\boldsymbol{\theta}\mid I)
&= p(S\mid I)\;\sum_{m,z} p(\boldsymbol{\theta}\mid m,z)\, p(m,z\mid I).
\end{split}
\label{eq:decomp}
\end{equation}
Therefore, to obtain a simulation-ready garment $G$ from $I$, we use a three-stage pipeline: (i) predict a garment geometry $G$ from $I$; (ii) infer \(p(M,Z\mid I)\) using abundant web-scale image-to-material supervision; and (iii) learn a material-to-physics mapper \(f_\phi:\,(M,Z)\!\mapsto\!\boldsymbol{\theta}\) from measured material datasets.
In practice, we implement
\begin{align}
\hat{S} &= \arg\max_{S} p(S\mid I),\\
(\hat{M},\hat{Z}) 
&= \arg\max_{m,z} p(m,z\mid I),\\
\hat{\boldsymbol{\theta}} 
&= f_\phi(\hat{M},\hat{Z}),
\label{eq:stages}
\end{align}
so that \(\hat{G} = (\hat{S},\hat{\boldsymbol{\theta}})\).
This split concentrates uncertainty into the first stage \(p(M,Z\mid I)\) (where rich supervision exists) and turns physics estimation into a well-posed supervised mapping \(p(\boldsymbol{\theta}\mid M,Z)\).

We use an existing image-to-garment geometry model to estimate $G$. To obtain $(m, z)$ and $\boldsymbol{\theta}$, we develop a custom estimation framework from in-the-wild images, leveraging our novel datasets. In the following, we first describe how we obtain these datasets and then how we train the corresponding models.

\subsection{Dataset Curation}

\paragraph{Fabric attributes from garment tags (FTAG) dataset.} We curate a garment material dataset for fabric composition understanding in apparel, containing \textit{16,026} images of people wearing garments paired with the corresponding material composition, fabric family, and fabric structure type annotations. The dataset was curated from publicly available online clothing stores and subsequently processed to filter out multi-layered garments and invalid data. Each entry corresponds to a single retail garment with vendor-provided material composition including exact fiber percentages, and a text description of the garment, from which we classify the fabric family and the fabric structure type. Our dataset reflects global fiber production trends~\cite{fernandez2021distribution,exchange2021preferred}, with a strong predominance of polyester and cotton, followed by viscose and nylon. We provide further details in the supplementary material.

\paragraph{Tag-to-Physics (T2P) dataset.} We curate a dataset to map intrinsic fabric properties to measurable physics parameters for simulation, containing \textit{1,254} fabrics paired with their corresponding fabric attributes and physical measurements. The dataset was compiled from publicly available online sources. Each entry corresponds to a single digitized fabric characterized by (1) fabric attributes, including multi-material composition $\mathcal{C} = \{(m_j, p_j)\}_{j=1}^{N}$, where $N$ is the number of constituent materials, $m_j$ denotes the $j$-th fiber type, and $p_j$ its corresponding percentage, fabric family $f$, structure type $s$, areal density $\rho$ (g/m$^2$), and thickness $t$ (mm); and (2) physics parameters, which quantify the mechanical response of fabrics under deformation. We denote the full set of physics parameters as: 
\begin{equation*}
    \mathcal{P} = \{\rho, p_{\text{friction}}, p_{\text{damping}}, p_{\text{ratio}}, p_{\text{stiffness}}\}
\end{equation*}
where $p_{\text{stiffness}}$ consists of buckling stiffness (g·mm$^2$/s$^2$),  
bending stiffness (g·mm$^2$/s$^2$),  
shear stiffness (g/s$^2$), and
stretch stiffness (g/s$^2$)\}. For anisotropic parameters, there are distinct values for the weft, warp, and bias directions. Each parameter describes a distinct aspect of a fabric’s mechanical behavior---governing resistance to compression, bending, in-plane shear, and tension---and are used commercial garment design software such as Blender~\cite{blender}, MarvelousDesigner~\cite{clo3d_website}, and Browzwear~\cite{browzwear}.  
These values are derived from standardized mechanical testing of textiles, providing a supervised mapping from interpretable fabric attributes to their measurable physical response.  
Fig.~\ref{fig:physics_parameter_dataset} in the supplementary visualizes the resulting distributions of fabric families, weave structures, material compositions, thickness, and areal density.
\subsection{Simulation-ready Garment Pipeline}
\label{sec:method}

Our single-image to simulation-ready garment generation pipeline consists of three stages. 
First, we reconstruct the garment geometry by estimating its sewing pattern using ChatGarment~\cite{bian2025chatgarmentgarmentestimationgeneration}. Because ChatGarment is trained on a diverse set of body poses to estimate sewing patterns draped on a standard A-pose human body, we can recover the garment geometry by draping the pattern with a standard cloth simulator~\cite{GarmentCode2023}. 
Next, we predict the garment’s material composition and fabric attributes, including fabric family, structure type, areal density, and thickness. 
Finally, we map these descriptors to the physical parameters required by the simulator using a learned material-to-physics model. 
An overview of the full pipeline is shown in Fig.~\ref{fig:pipeline}. 
In the remainder of this section, we focus on our main technical contribution: predicting fabric attributes and simulator-compatible physical parameters from a single image.

\subsection{Fabric Attribute Prediction}
\begin{table*}[t]
\centering
\caption{\textbf{Quantitative comparison of image-to-garment prediction.} We show results for geometry (CD, IoU), image-space-reconstruction (PSNR, SSIM, LPIPS), and physics parameter accuracy (NMAE) with respect to the ground truth of our created data. Best results in \textbf{bold}. Arrows indicate optimization direction.}
\label{tab:sup_quant}
\begin{tabular}{llcccccc}
\toprule
\textbf{Sequence} & \textbf{Method} & \textbf{\#Frames} & \textbf{CD}$\downarrow$ & \textbf{IoU}$\uparrow$ & \textbf{PSNR}$\uparrow$ & \textbf{SSIM}$\uparrow$ & \textbf{LPIPS}$\downarrow$\\
\midrule
\multirow{4}{*}{Jumping Jack} 
& GarmentRecovery* & \multirow{4}{*}{133}& 927.0 & 4.3 & 14.50 & 0.881 & 0.207 \\
& AIparrel* & & 98.1& 14.6 & 20.62 & 0.942 & 0.064\\
& ChatGarment* & & \underline{88.9} & \underline{20.0} & \underline{21.66} & \underline{0.951} & \underline{0.056}\\
& Image2Garment~(ours) &  & \textbf{64.4}& \textbf{21.6} & \textbf{22.10} & \textbf{0.954} & \textbf{0.053} \\
\midrule
\multirow{4}{*}{Joyful Jump} 
& GarmentRecovery* & \multirow{4}{*}{91}  & 58.7 & 11.7 & 18.46 & 0.934 & 0.102 \\
& AIparrel* & 
& 46.8 & 17.4 & 24.64 & \underline{0.965} & \underline{0.034} \\
& ChatGarment* & 
& \underline{7.7} & \underline{37.5} & \underline{27.88} & \textbf{0.970} & \textbf{0.021} \\
& Image2Garment~(ours)  & 
& \textbf{7.5} & \textbf{38.6} & \textbf{28.05} & \textbf{0.970} & \textbf{0.021} \\

\midrule
\multirow{4}{*}{Northern Spin} 
& GarmentRecovery* &\multirow{4}{*}{125}  & 275.0 & 5.3 & 14.00 & 0.828 & 0.262 \\
& AIparrel* & & \underline{257.0}& \textbf{14.0} & 17.86 & \textbf{0.881} & 0.156 \\
& ChatGarment* & & 525.0 & \underline{13.9} & \underline{18.32} & \underline{0.879} & \underline{0.150} \\
& Image2Garment~(ours)  & & \textbf{163.0} & 12.4 & \textbf{18.91} & 0.866 & \textbf{0.145}\\

\midrule
\multirow{4}{*}{Hit Reaction} 
& GarmentRecovery* &\multirow{4}{*}{62}  & 250.0 & 7.2 & 20.67 & 0.957 & 0.083 \\
& AIparrel* & & \textbf{57.6} & 19.3 & 20.59 & \underline{0.958} & 0.056 \\
& ChatGarment* & & 109.0 & \underline{20.6} & \underline{21.62} & 0.957  & \underline{0.056}  \\
& Image2Garment~(ours)  & & \underline{101.0} & \textbf{23.0} & \textbf{22.10}  & \textbf{0.961}  & \textbf{0.049} \\
\midrule

\multirow{4}{*}{\textbf{Average}} 
& GarmentRecovery* &\multirow{4}{*}{103} & 377.7 & 7.2 & 16.91 & 0.900 & 0.164 \\
& AIparrel* & & \underline{114.9} & 16.3 & 20.93 & 0.937 & 0.078\\
& ChatGarment* & & 182.7 & \underline{23.0} & \underline{22.37} & \textbf{0.939} & \underline{0.071} \\
& Image2Garment~(ours)  & & \textbf{84.0} & \textbf{23.9} & \textbf{22.79} & \underline{0.938} & \textbf{0.067}\\
\bottomrule
\end{tabular}
\end{table*}
In this step, we estimate the material composition $M$ and garment attributes $Z$ from a single image, where $M$ represents the fiber-level material mixture and $Z$ includes attributes such as fabric family and structure type. A key observation is that these attributes differ significantly in how identifiable they are from visual cues alone. In particular, the conditional distributions $p(m \mid I)$ can be highly multimodal: many distinct material compositions produce nearly indistinguishable appearances in photographs due to similar textures, colors, and weave patterns. By contrast, higher-level descriptors such as fabric family or structure type tend to exhibit clearer visual signatures and therefore admit a more peaked distribution. Attributes like areal density or thickness, however, often lack direct visual indicators and are therefore the most ambiguous.
We use this insight to motivate the model design in the following.

\paragraph{Material estimation model.}
These observations lead us to estimate the material composition with a generative model which not only already possesses general knowledge regarding fabric materials and garments but also supports multi-modal inputs that are essential for material prediction. In particular, we formulate the prediction of the fabric attributes as an image captioning task. For this, we prompt an existing VLM with the unaltered image and a custom text prompt and ask it to output a structured JSON string containing per-fiber percentages and other attributes. In the JSON string, we ask the VLM to include the following fabric properties for each input garment image that are essential for downstream physical parameter estimation:

\begin{itemize}
    \item Material composition $\hat{C}=\{(\hat{m}_j,\hat{p}_j)\}_{j=1}^{\hat{N}}$ denotes that the garment contains  $\hat{p}_j$ percent of fabric $\hat{m}_j \in \mathcal{M}$, a fixed set of apparel fibers 
    \footnote{E.g., cotton, polyester, elastane.}. This representation captures the 
    fiber-level makeup of the fabric, which directly governs its mechanical behavior.

    \item Fabric family $\hat{f}\in\mathcal{F}$, where $\mathcal{F}$ is a set of common 
    textile families\footnote{See the Supplementary Material for a full list.} such as denim, chiffon, or jersey, describes the fabric’s microstructure, drape, and surface appearance.

    \item Structure type $\hat{s}\in\{\text{knit}, \text{woven}, \text{others}\}$ identifies 
    the fabric construction. This structural prior restricts the physically plausible range of density, thickness, and stiffness parameters used in downstream simulation.
\end{itemize}

We use Qwen-2.5VL~\cite{bai2025qwen25vltechnicalreport} as our base VLM model and fine-tune it with LoRA~\cite{hu2022lora}. Because there is an imbalance in the frequency of materials in the FTAG dataset, we counteract this by specifying
weights $w_{t_i}$ as the inverse of the class frequency of the corresponding material during cross-entropy loss computation. Specifically, the model is fine-tuned using a weighted token-level cross-entropy loss:
\[
\mathcal{L}_{\text{VLM}}=-\sum_{i} w_{t_i}\log p_\theta(t_i \mid \mathcal{I},\mathbf{t}_{<i}),
\]
where $\mathbf{t}$ is the serialized target sequence.

\paragraph{Density–thickness estimator.}
Given a predicted fabric-attributes composition $\hat{C}=\{(\hat{m}_j,\hat{p}_j)\}_{j=1}^{\hat{N}}$, fabric family $\hat{f}$, and structure type $\hat{s}$---we estimate areal density $\hat{\rho}$ and thickness $\hat{t}$ by sampling from our fabric dataset:
\[
\mathcal{D}=\{(C_i,f_i,s_i,\rho_i,t_i)\}_{i=1}^{M}, \quad C_i=\{(m_{ij},p_{ij})\}_{j=1}^{N_i}.
\]
We perform a hierarchical search among fabrics sharing the same $(f_i,s_i)$:
(1) exact composition match,
(2) exact material type match, and
(3) primary material match. 
If multiple candidates exist, $\hat{\rho}$ and $\hat{t}$ are jointly sampled from the same fabric entry within the most specific non-empty set using one of three modes: mean, median, or random. We train using a $70 / 15 / 15$ split (train / validation / test) of our fabric dataset and use 5-fold stratified cross-validation to select the sampling mode with the lowest validation MAE. This hierarchical retrieval preserves the empirical correlation between density and thickness while ensuring robustness when exact composition matches are unavailable.

\subsection{Physics Parameter Prediction}
We predict physics parameters $\mathcal{P}$ from fabric attributes $(C,f,s,\rho,t)$ using independent Random Forest Regressors (RFRs)~\cite{randomforest} for bending-, shear-, stretch-, buckling-stiffness, and buckling ratio.  
Each Random Forest regressor is trained on ground-truth fabric attributes and corresponding physics parameters, following \cite{dominguezelvira2024MechFromMet}. We train using a $70 / 15 / 15$ split (train / validation / test) of our fabric dataset and tune hyper-parameters via a 50-iteration randomized search with 5-fold stratified cross-validation, selecting the configuration that minimizes validation MAE. During inference, we use predicted attributes $(\hat{C},\hat{f},\hat{s},\hat{\rho},\hat{t})$ as input to the same models. Friction and internal damping are fixed constants, while $\rho$ and $t$ are directly taken from predicted fabric attributes. 

\begin{figure*}[th]
    \centering
    \includegraphics[width=\textwidth]{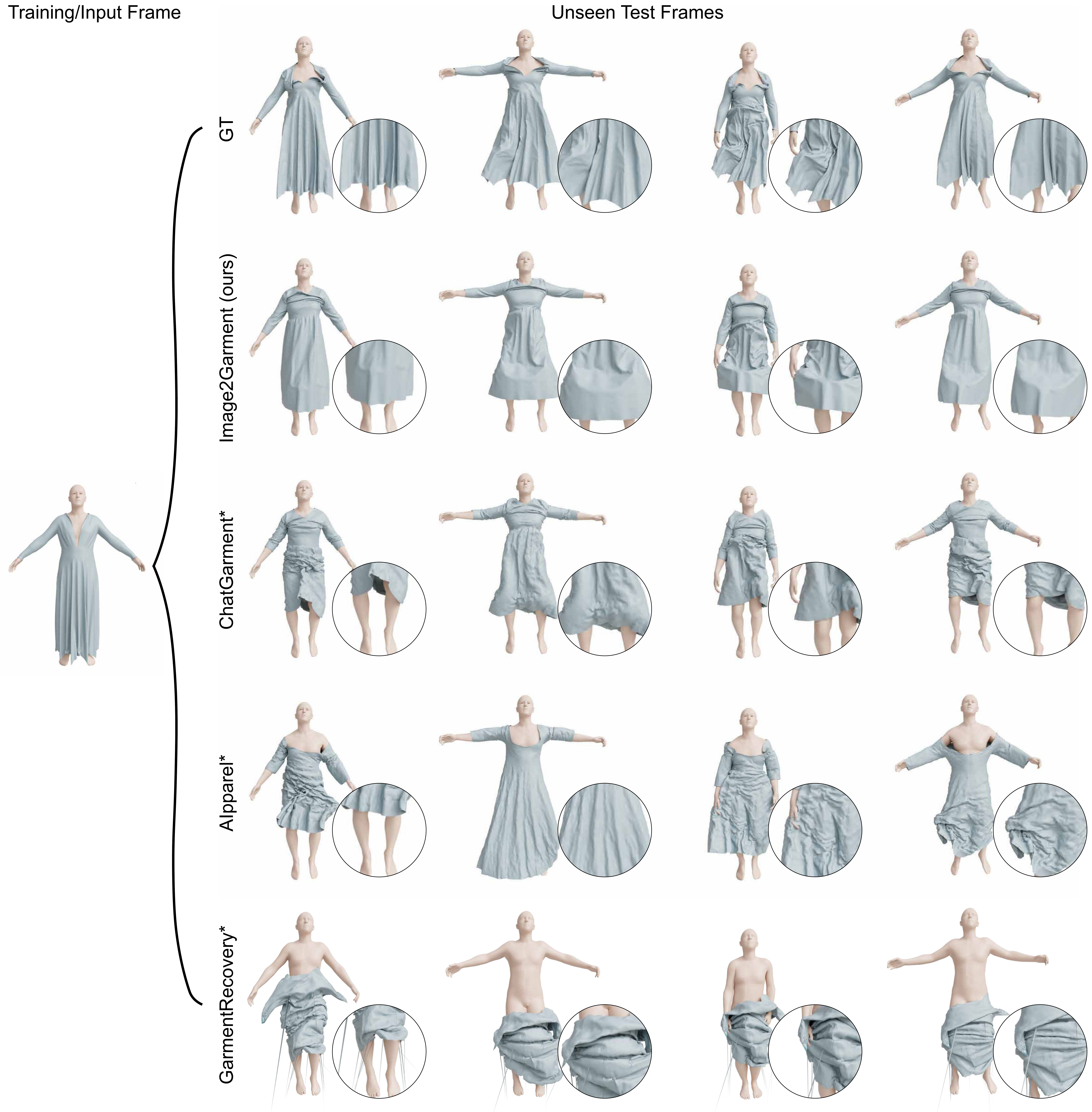}
    \caption{\textbf{Qualitative comparison on the jumping jack example.} The left most column shows the input frame presented to each method and the columns on the right show the animated and rendered garments at different points in time of the animation. Our method's result most closely resemble the ground truth (top row).}
    \label{fig:sup_qual_1}
    \vspace{-12pt}
\end{figure*}

\section{Experiments}
\vspace{-10pt}
\label{sec:experiments}
\paragraph{Training configuration.}
We fine-tune Qwen-2.5VL~\cite{bai2025qwen25vltechnicalreport} on the FTAG dataset (Sec.~\ref{sec:method}) using its standard train/val/test split of 9,843/1,231/1,231 samples. To address class imbalance in material composition, we apply inverse-frequency token weights. The model is trained for 5,200 iterations with a global batch size of 9, using LoRA~\cite{hu2022lora} (rank $R=64$) applied to all modules except [\texttt{lm\_head}, \texttt{embed\_tokens}]. We optimize with AdamW (learning rate $1\times10^{-5}$, weight decay $0.1$, $(\beta_1,\beta_2)=(0.9,0.999)$) and a cosine schedule. Images are dynamically resized between $1500\!\times\!1500$ and $1700\!\times\!1700$ pixels using the Qwen2.5-VL image processor. Training runs on 9 Quadro RTX 8000 GPUs and takes approximately 80 hours~\cite{kapfer2025marlowe}.

\begin{figure*}[th]
    \centering
    \includegraphics[width=\textwidth]{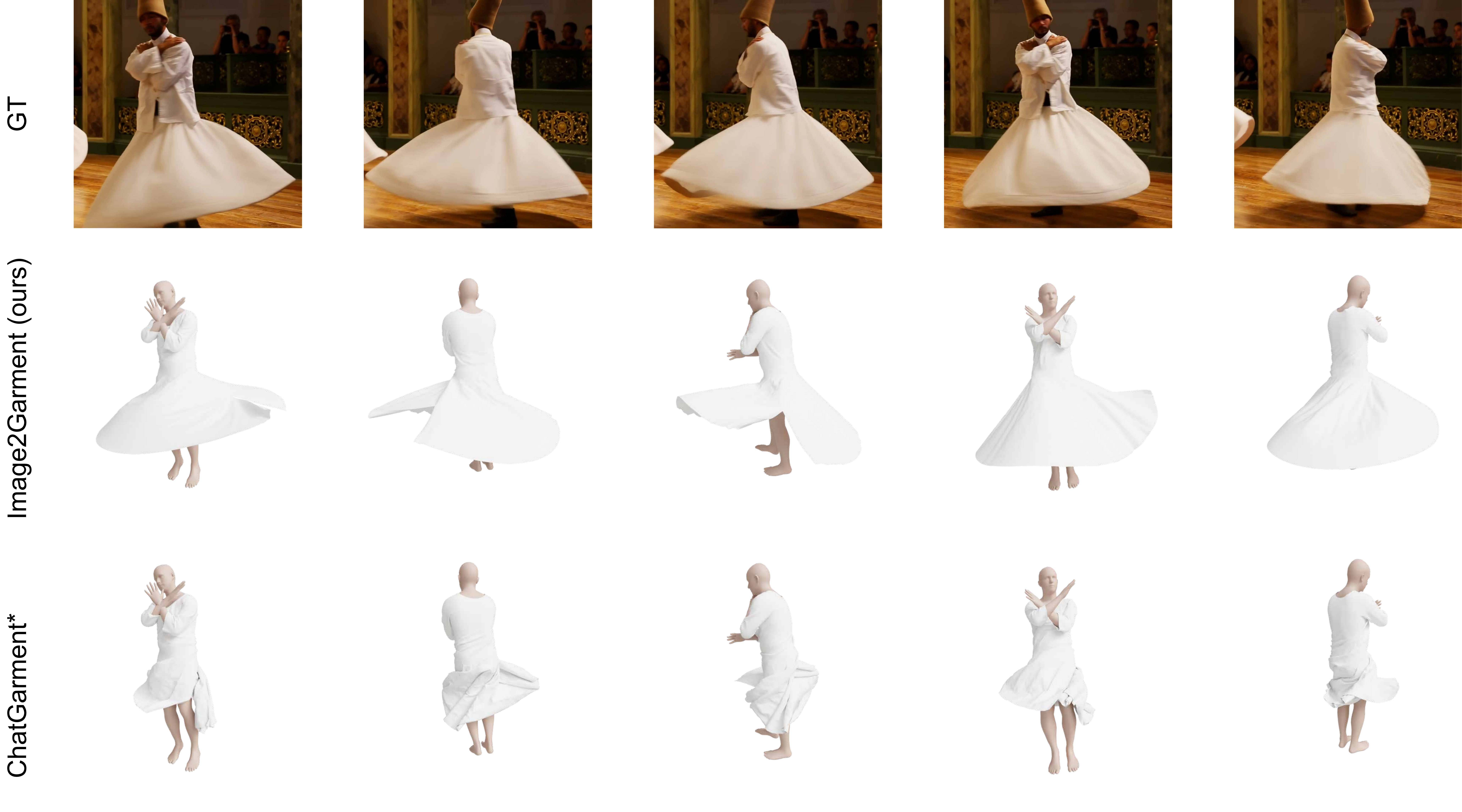}
    \caption{\textbf{Qualitative comparison on in-the-wild video.} The top row shows the original sequence. We only use the left-most frame as input. The rows below show renderings of the garments after simulation.}
    \label{fig:itw2}
    \vspace{-10pt}
\end{figure*}

\paragraph{Simulation setup.}
All garment simulations are performed in the Marvelous Designer (CLO3D)~\cite{MarvelousDesigner} engine, using its standard cloth solver and parameter definitions. We import the draped shapes together with our predicted physics parameters and simulate garments on the SMPL body for each sequence at 24\,fps. Cloth resolution is set to 20\,mm particle distance, with timestep 0.042\,s, gravity $-9800$\,cm/s$^2$, and default friction and damping settings (air damping $1.0$). Self-collision and body--cloth collision are enabled using iteration-based collision detection (50 CG iterations). All methods are evaluated under identical simulator settings to ensure that differences in dynamic behavior arise solely from the predicted geometry and physics parameters.
\vspace{-10pt}

\begin{table*}[t]
\centering
\caption{\textbf{Ablation of the Physics Parameter Prediction.} Geometry (CD, IoU), image reconstruction (PSNR, SSIM, LPIPS), and physics parameter accuracy (NMAE) with respect to ground truth. Our method estimates physical parameters through a feedforward method, while Random Parameters are sampled within the bounds of the simulator. Best results in \textbf{bold}. Arrows indicate optimization direction.}
\label{tab:sup_ablations}
\begin{tabular}{llcccccc}
\toprule
\textbf{Sequence} & \textbf{Method} & \textbf{\#Frames} & \textbf{CD}$\downarrow$ & \textbf{IoU}$\uparrow$ & \textbf{PSNR}$\uparrow$ & \textbf{SSIM}$\uparrow$ & \textbf{LPIPS}$\downarrow$ \\
\midrule
\multirow{2}{*}{Jab Cross} 
& Random Parameters & \multirow{2}{*}{105}& 5.42& 46.1 & 29.04 & 0.968 & 0.020\\
& Image2Garment~(ours)  &  & \textbf{2.50}& \textbf{58.4} & \textbf{30.99} & \textbf{0.974} & \textbf{0.014}\\
\midrule
\multirow{2}{*}{Jumping Jack} 
& Random Parameters & \multirow{2}{*}{133}& 79.4& 22.2 & 22.14 & 0.951 & 0.053 \\
& Image2Garment~(ours)  &  & \textbf{31.2}& \textbf{28.3} & \textbf{24.63} & \textbf{0.960} & \textbf{0.035} \\
\midrule
\multirow{2}{*}{Northern Spin} 
& Random Parameters & \multirow{2}{*}{150}& 68.6& 25.1 & 23.92 & 0.932 & 0.063 \\
& Image2Garment~(ours)  &  & \textbf{1.35} & \textbf{76.4} & \textbf{33.60} & \textbf{0.970} & \textbf{0.009} \\
\midrule
\multirow{2}{*}{Hit Reaction} 
& Random Parameters & \multirow{2}{*}{62}& 44.2 & 33.8 & 27.12 & 0.976 & 0.026 \\
& Image2Garment~(ours)  &  & \textbf{8.78} & \textbf{43.6} & \textbf{30.71} & \textbf{0.983} & \textbf{0.013} \\
\midrule
\multirow{2}{*}{\textbf{Average}} 
& Random Parameters & \multirow{2}{*}{88} & 49.41 & 31.80 & 25.56 & 0.957 & 0.041  \\
& Ours &  & \textbf{10.96} & \textbf{51.68} & \textbf{29.98} & \textbf{0.972} & \textbf{0.018}\\
\bottomrule
\vspace{-10pt}
\end{tabular}
\end{table*}
\paragraph{Evaluation dataset.} To compare our method further with exiting state-of-the-art methods we create the first synthetic evaluation dataset designed with complete garment prediction in mind. In particular our dataset contains four outfits sampled from~\cite{GarmentCodeData:2024} that each have been assigned a single realistic material per garment. We drape this garment over SMPL~\cite{SMPL:2015} bodies and create distinct animations using Mixamo~\cite{Mixamo}. We then use CLO3D's~\cite{clo3d_website} simulator to simulate the garment under the movement of the animation. Finally, we render the garment at each animation step using Blender~\cite{blender}.
From this data, we use the first rendered frame as an input and the remaining frames and meshes for evaluation.
Compared to existing evaluation benchmarks, this data has the advantage of including more complex garments and animations and does not suffer from measurement inaccuracies.

\vspace{-10pt}
\paragraph{Metrics.}
We quantitatively evaluate our pipeline using both shape-level and image-level metrics at each simulation timestep.
For shape-level evaluation, we compute the Chamfer Distance (CD) between point clouds uniformly sampled from the predicted and ground-truth garment surfaces. In addition, we measure volumetric Intersection-over-Union (IoU) by voxelizing both meshes at a fixed resolution and computing the overlap between occupied voxels at each timestep.
For image-level evaluation, we render the predicted and ground-truth garments from identical camera viewpoints at each timestep. We then apply standard image-space metrics commonly used in 3D reconstruction, including PSNR, SSIM, and LPIPS, to quantify perceptual and photometric discrepancies between the rendered images.

\vspace{-10pt}
\paragraph{Baselines.}
We compare our results against state-of-the-art garment shape estimation methods. In particular, we compare against ChatGarment~\cite{bian2025chatgarmentgarmentestimationgeneration}, AIpparel~\cite{nakayama2025aipparel}, and GarmentRecovery~\cite{li2024garment}. As none of these methods produce physical parameters for simulation, we augment these baselines by attaching a random material. We guarantee physical plausibility by sampling within the simulator bounds and mark the resulting methods with a *.

\paragraph{Quantitative evaluation.}
In this section, we evaluate our method on the dataset we created and compare it against state-of-the-art garment shape estimators paired with random material samples.
\cref{tab:sup_quant} reports the quantitative results. Our method achieves the best average performance across all 3D metrics and most 2D metrics. The only exception is SSIM, where our score is slightly lower than ChatGarment*; however, this can be attributed to the uniformly high SSIM values achieved by all methods, which reduces the discriminative power of this metric.
\vspace{-5pt}

\paragraph{Qualitative evaluation.}
Here, we present qualitative comparisons between our method and state-of-the-art garment generation approaches using random physics parameters.
\cref{fig:sup_qual_1} shows results for the \"jumping jack\" sequence in our dataset. While all the baselines struggle with either the shape of the garment or its deformation under simulated forces, we can observe that our method generates plausible geometry and physics parameters that remain
consistent with the ground truth throughout the animation. 
Further, we compare our performance with ChatGarment on some in-the-wild data from web videos. \cref{fig:itw2} shows the result of this experiment. We observe that the dress generated by our method aligns more closely with the ground-truth garment. The random parameters in ChatGarment cause the dress not to lift far enough. The remaining benchmark examples and more in-the-wild generations are visualized in the supplementary material.
\vspace{-5pt}

\paragraph{Ablations.} 
We conduct ablations at the 4D shape level using our curated dataset. To isolate the effect of physics parameter prediction, we use the ground-truth meshes and vary only the physics parameters. The quantitative results are reported in \cref{tab:sup_ablations}. Our predicted physics parameters yield substantial improvements across all metrics and sequences. Although both our method and the baseline exhibit some variance, our approach consistently outperforms the baseline overall, highlighting the importance of accurate physics parameter estimation. We further provide ablations of the material estimation in the supplementary material.
\section{Conclusion}
\label{sec:conclusion}
We introduce a feed-forward pipeline that recovers simulation-ready garments from a single image by first predicting structured, interpretable fabric attributes and then mapping them to physically valid simulator parameters. This factorized formulation transforms an otherwise ill-posed inverse problem into two learnable and data-efficient components. Together with our FTAG and T2P datasets, it enables accurate estimation of material composition, fabric family, structure type, density, thickness, and full physics parameters directly from in-the-wild images. Extensive quantitative and qualitative results show that our method achieves state-of-the-art performance in fabric attribute estimation, physics prediction, and dynamic draping quality, while being significantly faster and more scalable than optimization-based approaches. 

Although our model performs well across diverse garments, it is currently limited to single-layer clothing and may miss subtle material cues not fully observable from a single image. Our FTAG dataset does contain multi-layered garments which we filtered out for this work but could enable future research on complex, layered garment analysis.  Addressing these limitations, including handling layered garments or incorporating richer visual signals, remains an interesting direction for future work. Overall, our results demonstrate that semantically grounded material attributes provide an effective foundation for accessible, high-fidelity image-to-simulation pipelines.

\section{Acknowledgements}
We would like to thank LVMH for partial support of this work. We acknowledge support from Kiyohiro Nakamaya’s NSF Graduate Research Fellowship and from JST ASPIRE under Grant No. JPMJAP2404. Finally, we thank Stanford's Marlowe cluster for providing GPU computing for model training and evaluation.

{
    \small
    \bibliographystyle{ieeenat_fullname}
    \bibliography{main}
}
\clearpage
\setcounter{page}{1}
\maketitlesupplementary

\begin{figure*}[t]
    \centering
    \includegraphics[width=\textwidth]{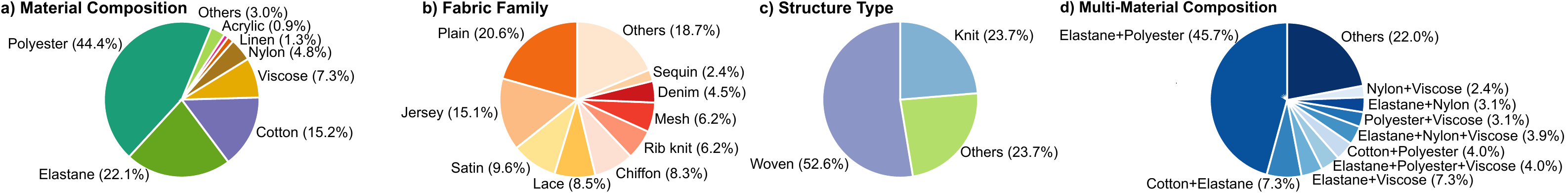}\\ 
    \caption{\textbf{Fabric attribute dataset.} 
Our Fabric Attributes from Garment Tags (FTAG) dataset contains 16,026 garment images paired with interpretable, vendor-provided fabric metadata. Each sample includes (i)~\emph{material composition}, specifying the fiber-level makeup that governs mechanical behavior; (ii)~\emph{fabric family}, characterizing microstructure, drape, and surface appearance; and (iii)~\emph{structure type}, indicating how yarns are interlaced or knitted to form the fabric. Subfigures show the distributions of (a)~main fiber compositions, (b)~fabric families, (c)~structure types, and (d)~multi-material fiber combinations.
}
\label{fig:ftag_dataset}
\end{figure*}

\section{Supplementary Website}

Please open the attached supplementary website to view our results in video, including dynamic draping simulations and qualitative comparisons.

\section{Overview}
This supplementary material provides additional details on our datasets,
model hyperparameters, and
evaluation protocols. We also include further quantitative and
qualitative results to complement the main paper.
Specifically, we cover:
\begin{itemize}
    \item Dataset curation and normalization for the Fabric Attributes from Garment Tags (FTAG) and Tag-to-Physics (T2P) datasets;
    \item Implementation details of the density--thickness estimator, and the physics parameter regressor;
    \item Detailed evaluation metrics and protocols for both fabric attribute prediction and dynamic draping;
    \item Additional quantitative comparisons and qualitative examples, including typical failure cases.
\end{itemize}

\section{Additional Dataset Details}

\begin{figure*}[t]
    \centering
    \includegraphics[width=\textwidth]{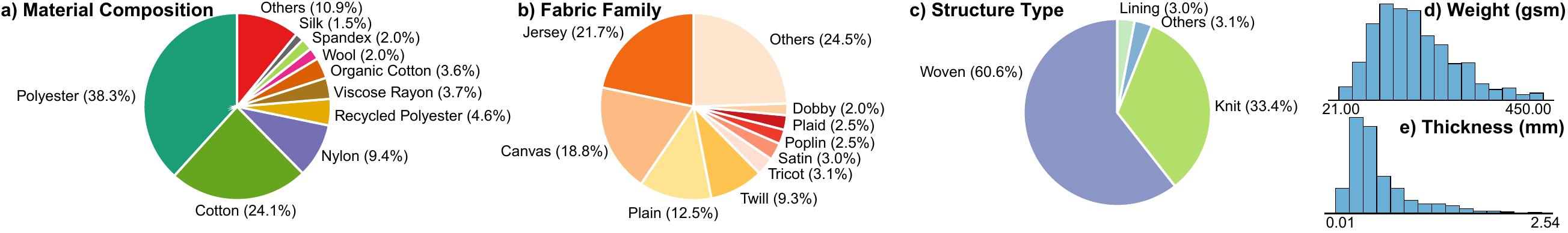}\\ 
    \caption{\textbf{Garment physics parameters dataset.} 
Our dataset contains 1,254 samples linking interpretable fabric attributes to measured fabric 
mechanical properties used for garment simulation. 
The fabric attributes include: material composition capturing fiber-level makeup that governs mechanical 
behavior; fabric family describing microstructure, drape, 
and surface appearance ; structure type identifying fabric construction that restricts the physically plausible range of simulation parameters; and density and thickness, which directly influence mass. 
Subfigures show distributions for (a)~main fiber compositions, (b)~fabric families, 
(c)~structure types, (d)~thickness, and (e)~weight.
}
    \label{fig:physics_parameter_dataset}
\end{figure*}

\subsection{FTAG: Fabric Attributes from Garment Tags}

\paragraph{Material composition estimation dataset.} We curate a garment material dataset for fabric attribute prediction,
containing \textit{16,026} images of models wearing retail garments paired
with three key labels: \emph{material composition}, \emph{fabric structure type},
and \emph{fabric family}. Images were collected from public online sources and processed through extensive cleaning, de-duplication, and normalization.

Each entry corresponds to a single retail garment with vendor-provided
\textit{material composition}
\[
C = \{(m_j, p_j)\}_{j=1}^{N},
\]
where each fiber type \( m_j \) belongs to a controlled vocabulary. We further classify the garment’s \textit{fabric structure type}:
\begin{small}
\begin{quote}
\emph{woven, knit, others}
\end{quote}
\end{small}
where woven fabrics correspond to interlaced yarns (e.g., denim, poplin, twill),
knit fabrics correspond to interlooped yarns (e.g., jersey, rib, fleece),
and others include non-woven categories such as mesh, lace,  and sequin fabrics.

Each garment also receives a fabric family label:
\begin{small}
\begin{quote}
\emph{chiffon, crepe, denim, satin, taffeta, poplin, twill, canvas, velvet, corduroy, flannel, organza, voile, lace, tulle, mesh, crochet, broderie, jacquard, sequin, metallic, leather, fleece, jersey, rib knit, french terry, interlock, pique, milano knit, ponte, scuba, tricot, pliss\'e, gingham, chambray, plain, cheesecloth, terry, unknown}
\end{quote}
\end{small}
selected from a controlled taxonomy based on visual texture, weave pattern,
drape, and surface appearance.

\paragraph{Processing and normalization details.} All garment entries come with vendor-provided material composition and fabric
descriptions that accompany each product image, allowing us to access standardized
fiber information and free-form fabric terminology. The material composition
is already reported in a consistent structured format (e.g., ``95\% Polyester,
5\% Elastane''), allowing us to extract fiber types directly and retain only
materials belonging to our validated vocabulary. During this step, we also
map synonymous fiber names to a canonical form (e.g., \emph{Spandex} and
\emph{Lycra Elastane} $\rightarrow$ \emph{Elastane}), ensuring that all
materials follow a unified naming scheme. Samples containing unrecognized
fibers or malformed percentage formats are discarded.

For the fabric family and fabric structure type, the vendor
descriptions include a wide variety of naming conventions (e.g., ``satin-style'',
``ribbed jersey'', ``denim-like weave''). To ensure consistency, we perform
an exhaustive search over these free-form descriptors and map each variant to
its canonical category in our controlled taxonomy. Examples include normalizing
``satin-style'' and ``sateen'' to \emph{satin}, ``ribbed knit'' to \emph{rib knit},
and ``mesh lace'' to \emph{mesh}.

The structure type is also provided in the vendor metadata and we
standardize it into the categories \emph{woven}, \emph{knit}, or \emph{others}.
We remove entries where the structure type contradicts the fabric family
(e.g., ``woven jersey'') or where the information is incomplete or ambiguous.

Our dataset also contains garments for which vendors report material composition
under multiple headers (e.g., \textit{Main}, \textit{Shell}, \textit{Lining},
\textit{Upper}), each corresponding to a different visible or internal layer
of the garment. Because our goal is to predict a single material composition
from a single image, we retain only entries whose composition is provided under
a primary header (\textit{Main} or other non-secondary headers) and discard
garments where the composition is split across multiple layers. This removes
cases such as jackets worn over shirts or dresses with separate lining materials,
where a single image cannot be associated with a unique fiber composition. The resulting FTAG dataset statistics and distributions are shown in Figure~\ref{fig:ftag_dataset}.

We additionally discard any sample for which one or more of the three attributes
---material composition, structure type, or fabric family---was missing, inconsistent,
or ambiguous. After applying these filtering steps, we obtain a final cleaned dataset
consisting of \textit{12{,}305} samples. We split this dataset in a stratified manner
based on the joint label of material composition, structure type, and fabric family,
resulting in \textit{9{,}843} training, \textit{1{,}231} validation, and
\textit{1{,}231} test samples for our fabric attribute estimation benchmark.

\subsection{Tag-to-Physics (T2P) Dataset}

We curate a dataset to convert intrinsic fabric properties to simulation
physics parameters, containing \textit{1,277} fabrics paired with
corresponding fabric attributes and CLO3D physics parameters.
The dataset is collected from publicly available online sources and
reflects real-world materials.

Each entry corresponds to a single digitized fabric with (1) fabric
attributes, which consist of:
\begin{itemize}
    \item \emph{Material composition} including exact fiber percentages,
    \emph{C $\in$ \{Acrylic, Alpaca, Angora, Cashmere, Cotton, Cupro, Hemp, Jute, Linen, Lyocell, Metallic, Modal, Nylon, Organic Cotton, PE (Polyethylene), PP (Polypropylene), PU (Polyurethane), PVC (Polyvinyl chloride), Polyester, Raccoon, Recycled Nylon, Recycled Polyester, Silk, Spandex, Supima Cotton, TENCEL\texttrademark, TPE (Thermoplastic Elastomer), TPU (Thermoplastic Poly Urethane), Triacetate, Viscose Rayon, Viscose from Bamboo, Wool, vinyl acetate copolymer)\}};
    \item \emph{Fabric family}
    \emph{f $\in$ \{Canvas, Challis, Chambray / Oxford, Chiffon, Circular Knit Spacer, Clip Jacquard, Corduroy, Crepe / CDC, Denim, Dobby, Dobby / Jacquard, Double Knit / Interlock, Double Weave, Flannel, Fleece, French Terry, Gauze / Double Gauze, Georgette, Interlining, Jacquard / Brocade, Jersey, Lace, Loop Terry, Memory, Mesh / Tulle, Organza, PVC, Pique, Plaid, Plain, Pointelle, Polar Fleece, Ponte, Poplin, Raschel, Rib, Ripstop, Satin, Seersucker, Sequin, Stretch Lining, Taffeta, Taffeta Lining, Tricot, Tweed, Twill, Vegan Fur, Vegan Leather, Vegan Suede, Velour, Velvet, Voile, Waffle\}};
    \item \emph{Structure type}
    \emph{s $\in$ \{Knit, Woven, Lining, Others\}};
    \item Areal density $\rho \in \mathbb{R}$ (g/m$^2$) and thickness $t \in \mathbb{R}$ (mm).
\end{itemize}

(2) CLO3D physics parameters quantify the mechanical response of fabrics
under deformation: \emph{P $\in$ \{$\rho$, friction, internal damping, buckling stiffness $\{b_{\text{bias-left}}, b_{\text{bias-right}}, b_{\text{warp}}, b_{\text{weft}}\}$, buckling ratio $\{r_{\text{bias-left}}, r_{\text{bias-right}}, r_{\text{warp}}, r_{\text{weft}}\}$, bending stiffness $\{B_{\text{bias-left}}, B_{\text{bias-right}}, B_{\text{warp}}, B_{\text{weft}}\}$ (g·mm$^2$/s$^2$), shear stiffness $\{S_{\text{left}}, S_{\text{right}}\}$ (g/s$^2$), and stretch stiffness $\{E_{\text{warp}}, E_{\text{weft}}\}$ (g/s$^2$)\}}. 

Each parameter describes a distinct aspect of the fabric’s mechanical behavior,
governing resistance to compression, bending, in-plane shear, and tension,
and is directly compatible with commercial garment design software such as
Marvelous Designer~\cite{MarvelousDesigner}, Browzwear~\cite{browzwear}, CLO3D, and proprietary ones such as Seddi Author \cite{dominguezelvira2024MechFromMet}. Each fabric sample is measured using the CLO
Fabric Kit; we use these measurements without additional scale factors, enabling
a supervised mapping from interpretable fabric attributes to measurable mechanical
response. Figure~\ref{fig:physics_parameter_dataset} in the main paper visualizes
the distributions of fabric families, structure types, material compositions,
thickness, and areal density.

\subsection{Material Prediction Metrics}

We report two primary metrics for evaluating material prediction performance:

\textbf{Average Accuracy:}
\begin{equation}
\text{Accuracy} = \frac{1}{N} \sum_{i=1}^{N} \frac{\text{TP}_i}{\text{TP}_i + \text{FP}_i + \text{FN}_i}
\end{equation}

Average accuracy across all examples, where for each example $i$, we compute the ratio of correctly predicted materials (true positives) to the total number of unique materials in either the ground truth or predictions. $N$ is the total number of examples.

\textbf{Average F1 Score:}
\begin{equation}
\text{F1} = \frac{1}{N} \sum_{i=1}^{N} \frac{2 \times \text{Precision}_i \times \text{Recall}_i}{\text{Precision}_i + \text{Recall}_i}
\end{equation}

where $\text{Precision}_i = \frac{\text{TP}_i}{\text{TP}_i + \text{FP}_i}$ and $\text{Recall}_i = \frac{\text{TP}_i}{\text{TP}_i + \text{FN}_i}$.

Average F1 score across all examples, computed as the harmonic mean of precision and recall for each example. For each example $i$, $\text{TP}_i$ represents correctly identified materials, $\text{FP}_i$ represents incorrectly predicted materials, and $\text{FN}_i$ represents ground truth materials that were missed.

\subsection{Material Percentage Prediction Metrics}

We report two metrics for evaluating the accuracy of predicted material percentages:

\textbf{Average Mean Absolute Error (MAE):}
\begin{equation}
\text{MAE} = \frac{1}{N} \sum_{i=1}^{N} \frac{1}{|M_i|} \sum_{m \in M_i} |p_{i,m}^{\text{gt}} - p_{i,m}^{\text{pred}}|
\end{equation}

Average mean absolute error across all examples, where for each example $i$, we compute the absolute difference between ground truth percentage $p_{i,m}^{\text{gt}}$ and predicted percentage $p_{i,m}^{\text{pred}}$ for each material $m$ in the set $M_i$ (all unique materials in either ground truth or predictions for example $i$). $|M_i|$ is the total number of materials for example $i$, and $N$ is the total number of examples.

\textbf{Average Normalized Mean Absolute Error (NMAE):}
\begin{equation}
\text{NMAE} = \frac{1}{N} \sum_{i=1}^{N} \frac{1}{|M_i|} \sum_{m \in M_i} \frac{|p_{i,m}^{\text{gt}} - p_{i,m}^{\text{pred}}|}{\max(p_{i,m}^{\text{gt}}, p_{i,m}^{\text{pred}})}
\end{equation}

Average normalized mean absolute error across all examples, where the absolute error for each material is normalized by the maximum of the ground truth and predicted percentages to account for varying material proportions. This provides a scale-invariant measure of percentage prediction accuracy.
\subsection{Structure Type Prediction Metrics}
We report two metrics for evaluating fabric structure type classification:

\textbf{Overall Accuracy:}
\begin{equation}
\text{Accuracy} = \frac{1}{N} \sum_{i=1}^{N} \mathbbm{1}(s_i^{\text{gt}} = s_i^{\text{pred}})
\end{equation}

Overall accuracy across all examples, where $s_i^{\text{gt}}$ is the ground truth structure type for example $i$, $s_i^{\text{pred}}$ is the predicted structure type, and $\mathbbm{1}(\cdot)$ is the indicator function that equals 1 when the condition is true and 0 otherwise. $N$ is the total number of examples.

\textbf{Macro-Averaged F1 Score:}
\begin{equation}
\text{F1}_{\text{macro}} = \frac{1}{|\mathcal{S}|} \sum_{s \in \mathcal{S}} \text{F1}_s
\end{equation}

where 
\begin{equation}
\text{F1}_s = \frac{2 \times \text{Precision}_s \times \text{Recall}_s}{\text{Precision}_s + \text{Recall}_s}
\end{equation}

and $\text{Precision}_s = \frac{\text{TP}_s}{\text{TP}_s + \text{FP}_s}$, $\text{Recall}_s = \frac{\text{TP}_s}{\text{TP}_s + \text{FN}_s}$.

Macro-averaged F1 score computed by first calculating the F1 score for each structure type class $s \in \mathcal{S}$ (where $\mathcal{S}$ is the set of all structure types: \{knit, woven, others\}), then averaging across all classes. $\text{TP}_s$, $\text{FP}_s$, and $\text{FN}_s$ represent true positives, false positives, and false negatives for class $s$, respectively.

\subsection{Fabric Family Prediction Metrics}
We report two metrics for evaluating fabric family classification:

\textbf{Overall Accuracy:}
\begin{equation}
\text{Accuracy} = \frac{1}{N} \sum_{i=1}^{N} \mathbbm{1}(f_i^{\text{gt}} = f_i^{\text{pred}})
\end{equation}

Overall accuracy across all examples, where $f_i^{\text{gt}}$ is the ground truth fabric family for example $i$, $f_i^{\text{pred}}$ is the predicted fabric family, and $\mathbbm{1}(\cdot)$ is the indicator function that equals 1 when the condition is true and 0 otherwise. $N$ is the total number of examples.

\textbf{Macro-Averaged F1 Score:}
\begin{equation}
\text{F1}_{\text{macro}} = \frac{1}{|\mathcal{F}|} \sum_{f \in \mathcal{F}} \text{F1}_f
\end{equation}

where 
\begin{equation}
\text{F1}_f = \frac{2 \times \text{Precision}_f \times \text{Recall}_f}{\text{Precision}_f + \text{Recall}_f}
\end{equation}

and $\text{Precision}_f = \frac{\text{TP}_f}{\text{TP}_f + \text{FP}_f}$, $\text{Recall}_f = \frac{\text{TP}_f}{\text{TP}_f + \text{FN}_f}$.

Macro-averaged F1 score computed by first calculating the F1 score for each fabric family class $f \in \mathcal{F}$ (where $\mathcal{F}$ is the set of all fabric families, including Poplin, Jersey, Lace, Satin, Chiffon, etc.), then averaging across all classes. $\text{TP}_f$, $\text{FP}_f$, and $\text{FN}_f$ represent true positives, false positives, and false negatives for class $f$, respectively.

\subsection{Density and Thickness Estimation Metrics}

We use our T2P dataset to estimate density and thickness values from ground truth fabric family, structure type, and material composition. We also prompt GPT to perform the same estimation for comparison (see Table~\ref{tab:metadata_estimation_results} in the main paper).

We report two metrics for evaluating continuous parameter estimation:

\textbf{Mean Absolute Error (MAE):}
\begin{equation}
\text{MAE} = \frac{1}{N} \sum_{i=1}^{N} |y_i^{\text{gt}} - y_i^{\text{pred}}|
\end{equation}

Mean absolute error between ground truth values $y_i^{\text{gt}}$ and predicted values $y_i^{\text{pred}}$ across all $N$ examples.

\textbf{Normalized Mean Absolute Error (NMAE):}
\begin{equation}
\text{NMAE} = \frac{\text{MAE}}{y_{\max} - y_{\min}}
\end{equation}

Normalized mean absolute error, where MAE is divided by the range of ground truth values ($y_{\max} - y_{\min}$) to provide a scale-invariant measure of prediction accuracy.

\subsection{Density--Thickness Estimator}
Given predicted discrete attributes
$(\hat{C}, \hat{f}, \hat{s})$ from the VLM, we estimate areal density $\hat{\rho}$
and thickness $\hat{t}$ using the T2P dataset
$\mathcal{D} = \{(C_i, f_i, s_i, \rho_i, t_i)\}_{i=1}^M$.
We perform hierarchical retrieval:
\begin{enumerate}
    \item \textit{Exact attribute match}:
    we first search for fabrics with $(f_i, s_i) = (\hat{f}, \hat{s})$ and
    a composition $C_i$ matching $\hat{C}$ exactly up to a small tolerance
    in percentages.
    \item \textit{Material-set match}:
    if no exact match exists, we relax to fabrics whose \emph{set} of fibers
    matches that of $\hat{C}$, ignoring small percentage differences.
    \item \textit{Primary fiber match}:
    if the previous sets are empty, we match fabrics whose primary fiber
    (highest percentage in $C_i$) equals the primary fiber in $\hat{C}$.
\end{enumerate}
From the most specific non-empty candidate set, we estimate $(\hat{\rho}, \hat{t})$
either by taking the mean, median, or by randomly sampling a single fabric entry.
We select the aggregation mode by performing 5-fold stratified cross-validation
on the T2P dataset (train:val:test = 70:15:15) and choosing the mode with lowest validation MAE.
The best performing mode was \textit{mean aggregation}.
This retrieval-based strategy preserves the empirical coupling between
areal density and thickness while remaining robust when exact composition
matches are not available.

\subsection{Physics Parameter Estimator}
We predict physics parameters $P$ from fabric attributes
$(C, f, s, \rho, t)$ using a collection of independent Random Forest
Regressors (RFRs), following the practical recipe of Dominguez-Elvira
\emph{et al.}~\cite{dominguezelvira2024MechFromMet}.
We form a feature vector by concatenating:
\begin{itemize}
    \item normalized fiber percentages over the vocabulary,
    \item one-hot encodings of fabric family $f$ and structure type $s$,
    \item scalar features $\rho$ and $t$ (optionally log-transformed).
\end{itemize}
We train five separate Random Forest Regressors, one for each physics parameter group. We use a 70/15/15 train/val/test split and perform a 50-iteration randomized hyperparameter search with 5-fold stratified cross-validation to select the number of trees and features, the max depth, and the minimum samples per split and leaf.

For training all models, we use Mean Absolute Error as the loss function. 
The best hyperparameters selected via cross-validation are: 
for bending, stretch, shear, and buckling stiffness, 
$n_{\text{estimators}}=100$, $\text{max\_depth}=20$, 
$\text{min\_samples\_split}=0.0420$, $\text{min\_samples\_leaf}=0.0094$, 
$\text{max\_features}=0.6911$; 
for buckling ratio, $n_{\text{estimators}}=200$, $\text{max\_depth}=30$, 
$\text{min\_samples\_split}=0.0703$, $\text{min\_samples\_leaf}=0.0016$, 
$\text{max\_features}=0.9595$.

During inference, we feed predicted
attributes $(\hat{C}, \hat{f}, \hat{s}, \hat{\rho}, \hat{t})$ into the same
regressors to obtain $\hat{P}$. Friction $\mu$ and internal damping $d$ are
kept fixed across fabrics, while $\hat{\rho}$ and $\hat{t}$ are directly
taken from the density--thickness estimator.

\section{Garment Geometry and Simulation Details}

\subsection{Sewing Pattern Estimation}

For garment geometry, we adopt ChatGarment as our image-to-sewing-pattern
backbone. Given a single image, ChatGarment predicts a sewing pattern
representation draped on a canonical SMPL body in A-pose. We export the
predicted pattern and re-simulate it using our cloth simulator, ensuring that
all methods (ours and baselines) are evaluated under identical numerical
simulation settings.

We use the default ChatGarment resolution and panel parameterization and
do not fine-tune ChatGarment on 4D-Dress to keep the comparison fair and
emphasize the benefits of improved physics prediction.

\section{Evaluation Protocols and Metrics}

\subsection{Dynamic Draping Metrics}

To evaluate dynamic draping on 4D-Dress, we compare our simulated garment
meshes against ground-truth cloth meshes over the entire sequence.
We use two metrics:

\begin{itemize}
    \item \textit{Chamfer Distance (CD) ↓:} We sample points from both the predicted and ground-truth garment meshes at each frame and compute the symmetric Chamfer distance. The reported CD is averaged over frames and scaled by $10^4$ for readability.
    \item \textit{Intersection-over-Union (IoU) ↑:} We voxelize each mesh
    using 5cm voxels and compute the intersection-over-union of the
    occupied voxels between predicted and ground-truth garments, then
    average over frames.
\end{itemize}

We report separate metrics for upper- and lower-body garments, as in
Table~1 of the main paper. Garments spanning upper and lower body, count towards both categories results.

\subsection{Fabric Attribute Prediction Metrics}

For fabric attribute estimation on FTAG, we evaluate:

\begin{itemize}
    \item \textit{Structure type} and \textit{fabric family}: macro-averaged
    accuracy and F1-score, computed by averaging per-class metrics across
    all categories.
    \item \textit{Material type}: multi-label accuracy and F1-score over
    the material vocabulary, ignoring true negatives to avoid inflated
    scores due to sparsity.
    \item \textit{Material percentages, density, and thickness}: mean
    absolute error (MAE) and normalized MAE (NMAE) over all test samples.
\end{itemize}

These metrics are reported for our fine-tuned Qwen2.5-VL model and for
ChatGPT baselines (zero- and few-shot) in the main paper.

\section{Additional Results}
\begin{figure*}[t]
\centering
\includegraphics[width=0.9\textwidth]{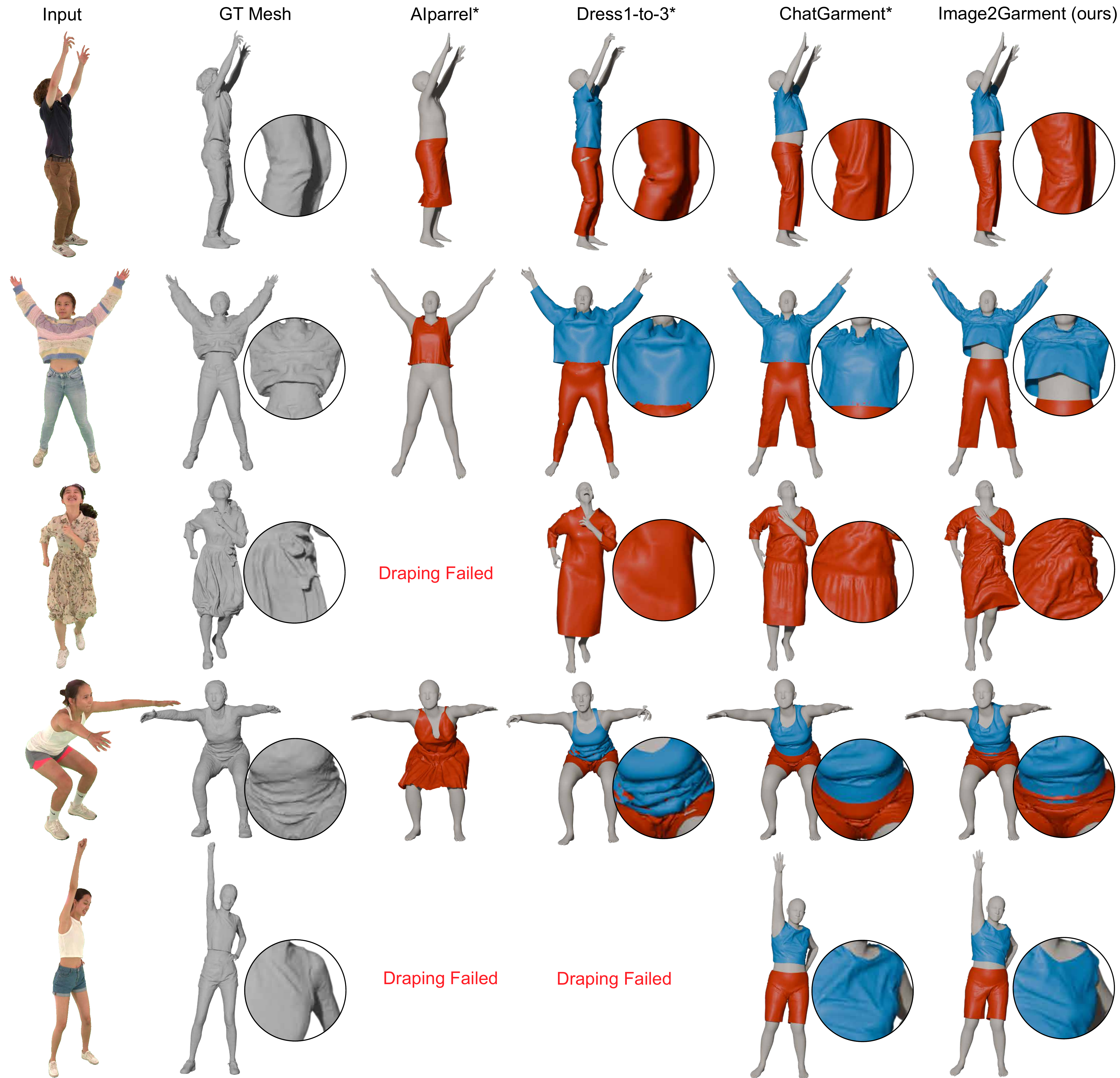}
\caption{
\textbf{Qualitative comparison of clothing reconstruction on 4D-Dress \cite{wang20244ddress}.}
We visualize reconstructed garment meshes from different approaches. The zoomed-in image next to the garment meshes highlights the differences resulting from runningthe  simulation.
Compared to baselines, our model produces garments with more realistic shapes and wrinkle patterns that are better aligned with the ground truth geometry. 
\textit{ChatGarment*} uses the same garment geometry as our method but randomly sampled physical parameters, leading to incorrect garment deformation and dynamics. 
}
\label{fig:4ddress_qualitative_comparison}
\end{figure*}

\subsection{Evaluation Datasets.}
\paragraph{4D-Dress dataset.}
In addition to the dataset we created synthetically, we also evaluate garment reconstruction and dynamic draping quality on the 4D-Dress dataset~\cite{wang20244ddress}. 
The dataset contains 50 multi-view video sequences of clothed subjects from which we select 5 sequences for evaluation.
For each sequence, we use the first camera view as the single-image input and use the provided cloth meshes for evaluation. 

\paragraph{FTAG dataset.}
For quantitative evaluation of material and fabric attribute prediction, we use the test split of our 
FTAG dataset, which includes 1,231 garments spanning 70 distinct material compositions 
(669 single-fiber and 562 multi-fiber blends), 28 fabric families, and 3 structure types, 
featuring challenging textures and diverse shapes.

\subsection{Baselines}
\paragraph{Single-image garment reconstruction.}
We compare against several state-of-the-art single-image garment reconstruction methods: 
ChatGarment~\cite{bian2025chatgarmentgarmentestimationgeneration}, AIpparel~\cite{nakayama2025aipparel}, and Dress-1-to-3~\cite{li2025dress}. 
We use the released checkpoints for ChatGarment and AIpparel. For Dress 1-to-3, we obtained the evaluating samples from the authors. 

\paragraph{Material and fabric attribute prediction.} We use OpenAI's GPT-5~\cite{openai_gpt5}, as a baseline as it is often credited with having the strongest visual recognition and reasoning capabilities. We test it in both a zero-shot and few-shot manner, providing it with the same information about plausible materials that we make available to our own model. 

\subsection{Evaluation Metrics}

\paragraph{Garment prediction.} For the simulation-ready garment prediction, we focus on the garment geometry across the entire simulation. We compare it to the ground truth using Chamfer Distance between the point clouds sampled from the predicted and ground-truth mesh, as well as the Intersection over Union (IoU) of the garment geometry. See the supplementary for details.

\paragraph{Fabric attribute prediction.} For structure type and fabric family, we report macro-averaged accuracy and F1-score, computed by averaging the per-class metrics across all categories. For material composition, we report per-example accuracy and F1-score averaged across all test examples. For material composition accuracy, we exclude true negatives from the calculation, as the large vocabulary of valid materials combined with the sparsity of material usage (garments typically contain 1-3 materials) would inflate accuracy scores artificially. For numerical values (material percentages), we compare predictions with ground truth using both the Mean Absolute Error (MAE) and Normalized Mean Absolute Error (NMAE).

\begin{table}[t]
\centering
\begin{tabular}{lcccc}
\toprule
& \multicolumn{2}{c}{\textbf{Lower}} & \multicolumn{2}{c}{\textbf{Upper}} \\
\cmidrule(lr){2-3} \cmidrule(lr){4-5}
\textbf{Method} & CD ↓ & IoU ↑ & CD ↓ & IoU ↑ \\
\midrule
Dress 1-to-3~\cite{li2025dress} & 34.1 & 44.8 & 40.5 & 40.0 \\
ChatGarment~\cite{bian2025chatgarmentgarmentestimationgeneration} & \underline{28.9} & \underline{48.4} & \underline{28.1} & \underline{46.6} \\
AIpparel~\cite{nakayama2025aipparel} & 380 & 28.0 & 380 & 28.0 \\
\midrule
\textbf{Ours} & \textbf{27.8} & \textbf{48.7} & \textbf{27.4} & \textbf{47.4} \\
\bottomrule
\end{tabular}
\caption{
\textbf{Clothing reconstruction benchmark.} 
We report the average Chamfer Distance (CD ↓) and Intersection over Union (IoU ↑) 
between the ground-truth garment meshes and the reconstructed clothing averaged over time
for the lower and upper garment categories in 4D-Dress. Our method achieves the best score in both metrics, showcasing the importance of accurate physical parameters for garment dynamics. Reported CD is multiplied by 1e4.
}
\label{tab:clothing_reconstruction}
\end{table}

\begin{table*}[t]
\centering
\begin{tabular}{lccc}
\toprule
\textbf{Attribute Field} & \textbf{ChatGPT (zero-shot)} & \textbf{ChatGPT (few-shot)} & \textbf{Ours} \\
\midrule
\multicolumn{4}{l}{\textit{Categorical Fields (Accuracy \% / F1-score)} ↑} \\
\midrule
Fabric Family & 0.58 / 0.42 & \underline{0.61} / \underline{0.43} & \textbf{0.75} / \textbf{0.72} \\
Structure Type & 0.74 / 0.68 & \underline{0.75} / \underline{0.69} & \textbf{0.86} / \textbf{0.85} \\
Material Type & 0.65 / \underline{0.70} & \underline{0.66} / \underline{0.70} & \textbf{0.71} / \textbf{0.75} \\
\midrule
\multicolumn{4}{l}{\textit{Continuous Fields (MAE \% / NMAE )} ↓} \\
\midrule
Material Percentage & 23.3 / 0.45 & \underline{22.4} / \underline{0.43} & \textbf{19.3} / \textbf{0.40} \\
Density (g/m$^{2}$) & \underline{64.38} / \underline{0.121} & 75.62 / 0.143 & \textbf{51.28} / \textbf{0.097} \\
Thickness (mm) & \underline{0.376} / \underline{0.053} & 0.378 / 0.054 & \textbf{0.227} / \textbf{0.032} \\
\bottomrule
\end{tabular}
\caption{
\textbf{Performance of fabric attribute estimation baselines.} 
ChatGPT (zero-shot) and ChatGPT (few-shot) results are compared with our finetuned Qwen2.5-VL model. 
Categorical fields are evaluated using Accuracy and F1 Score, while continuous-valued fields are evaluated using MAE and NMAE (\%). 
Weight and Thickness are estimated on the fabric test set using the thickness–density sampler (5-fold Cross Validation), 
and the remaining fields correspond to the stratified test split of our material composition estimation benchmark.
}
\label{tab:metadata_estimation_results}
\end{table*}

\subsection{Results}

\paragraph{Simulation-ready garment estimation.}
\cref{tab:clothing_reconstruction} reports quantitative results on single-image garment reconstruction and dynamic draping. Across all metrics, our approach outperforms existing state-of-the-art methods. In particular, our simulated garments achieve the lowest Chamfer Distance and highest IoU relative to ground-truth cloth geometry, demonstrating that both our reconstructed sewing patterns and our predicted physics parameters are functionally accurate across the entire simulation.

In addition to accuracy, our method is substantially more efficient than optimization-based approaches such as Dress-1-to-3, which require iterative fitting. In contrast, our feed-forward pipeline produces reconstruction and physics estimates in a single pass, enabling consistent performance on in-the-wild inputs without any human intervention. 

\cref{fig:4ddress_qualitative_comparison} provides a visual comparison on the 4D-Dress dataset. While all methods recover plausible static garment shapes, differences become pronounced under motion. Competing approaches frequently produce garments that over-stiffen, collapse, or drift away from the ground-truth dynamics. Our method, by contrast, generates garment motion and silhouettes that closely follow the ground-truth cloth over entire sequences. This highlights the importance of accurate fabric attribute and physics prediction: our inferred bending, shear, and stretch parameters yield dynamic deformations that better capture the true material behavior. 

\paragraph{Fabric attribute estimation.}
Table~\ref{tab:metadata_estimation_results} reports the performance of our fabric attribute estimation model compared to strong vision-language baselines. Across all fields, our fine-tuned Qwen2.5-VL model substantially outperforms both zero-shot and few-shot ChatGPT. For categorical attributes such as fabric family, structure type, and material type, our method yields markedly higher accuracy and F1-scores, with gains of up to 30\% absolute accuracy over zero-shot prompting. These improvements reflect the benefits of domain-specific finetuning and structured JSON supervision, which enable the model to resolve subtle appearance cues that general-purpose VLMs fail to capture. For continuous-valued fields, including material percentage, areal density, and thickness, our approach achieves the lowest MAE and NMAE across all metrics. Notably, thickness prediction error is reduced by nearly half compared to zero-shot ChatGPT, and weight estimation shows similar improvements. These results demonstrate that accurate fabric attribute prediction requires specialized training rather than generic prompting, and they establish our model as a reliable foundation for downstream physics parameter estimation and garment simulation.

\paragraph{Ablations.}
\begin{table}[t]
\centering
\caption{\textbf{Ablation study.} We evaluate the impact of fine-tuning, image cropping, majority-only material prediction, and joint prediction with fabric attributes on material composition, structure type, and fabric family estimation tasks.}
\label{tab:ablation}
\begin{tabular}{lcc}
\toprule
\textbf{Method Variant} & \textbf{Accuracy} $\uparrow$ & \textbf{F1-score} $\uparrow$ \\
\midrule
\multicolumn{3}{l}{\textit{Material Composition}} \\
Full Model & \textbf{71}\% & \textbf{0.75} \\
    \quad w/ Primary Fiber & 62\% & 0.68 \\
\quad w/o Fabric Attributes & \underline{69\%} & \underline{0.73} \\
\quad w/o Fine-tuning & 48\% & 0.56 \\
\quad w Cropped Images & 63\% & 0.67 \\
\midrule
\multicolumn{3}{l}{\textit{Structure Type}} \\
Full Model & \textbf{86}\% & \textbf{0.85} \\
\quad w/o Fine-tuning & 63\% & 0.33 \\
\quad w Cropped Images & \underline{75\%} & \underline{0.73} \\
\midrule
\multicolumn{3}{l}{\textit{Fabric Family}} \\
Full Model & \textbf{75}\% & \textbf{0.72} \\
\quad w/o Fine-tuning & 40\% & 0.31 \\
\quad w Cropped Images & \underline{58\%} & \underline{0.48} \\
\bottomrule
\vspace{-20px}
\end{tabular}
\end{table}

We conduct an ablation study to assess the importance of major components in our framework, as summarized in Table~\ref{tab:ablation}. Fine-tuning proves essential across all three tasks. Without fine-tuning, the zero-shot model’s performance drops by 23, 23, and 35 percentage points on material composition, structure type, and fabric family estimation, respectively, compared to our fully fine-tuned model. This substantial performance gap demonstrates that task-specific adaptation is critical for reliable fabric property estimation from visual data alone. For material composition estimation, predicting only the primary material type (w/ Primary Fiber) leads to a noticeable drop in both accuracy (71\% $\rightarrow$ 62\%) and F1-score (0.75 $\rightarrow$ 0.68), confirming that capturing the full material composition is critical.
Removing jointly predicted fabric attributes causes additional degradation (71\% $\rightarrow$ 69\% accuracy), indicating that these complementary cues provide useful supporting information for material composition estimation. Following~\cite{dominguezelvira2024MechFromMet}, we also evaluate using images of the fabrics only. To isolate the fabric region, we apply a 512$\times$512 center crop to each image. Center-cropping consistently harms performance across all tasks, with severe drops in material composition (71\% $\rightarrow$ 63\%), structure type (86\% $\rightarrow$ 75\%), and especially fabric family estimation (75\% $\rightarrow$ 58\% accuracy). This demonstrates that high-resolution detail and complete garment context---including silhouette, draping patterns, and peripheral regions---provide essential visual information for reliably inferring fabric properties. Overall, each component contributes meaningfully to final estimation quality, with fine-tuning and full-image context being the most critical factors.

\subsection{Additional Image-to-Garment Examples}
\begin{figure*}
    \centering
    \includegraphics[width=0.8\textwidth]{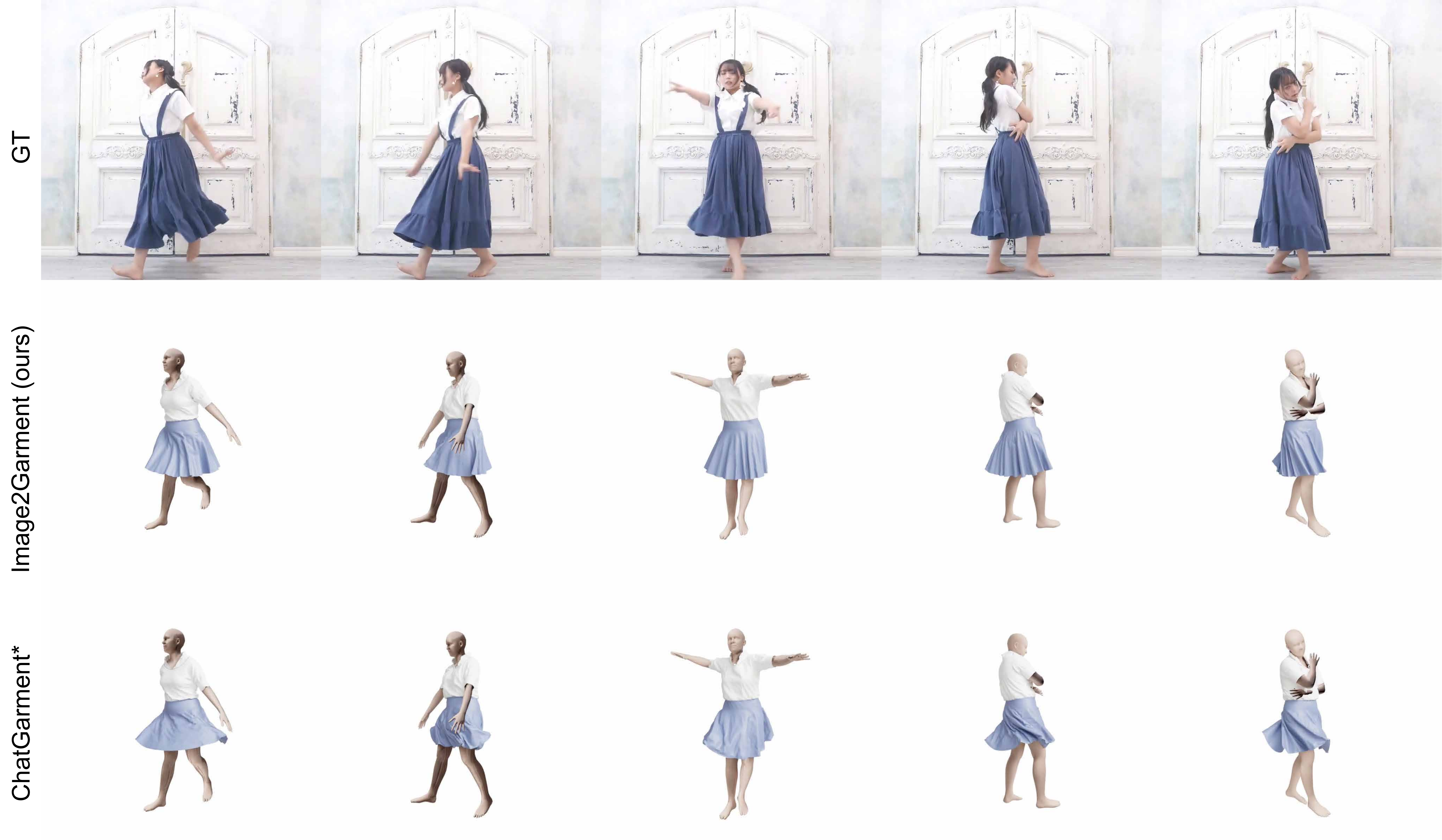}
    \caption{\textbf{Qualitative comparison in-the-wild video.} The top row shows the original sequence. We only use the left-most frame as input. The rows below show renderings of the garments after simulation.}
    \label{fig:itw1}
\end{figure*}

\begin{figure*}
    \centering
    \includegraphics[width=0.8\textwidth]{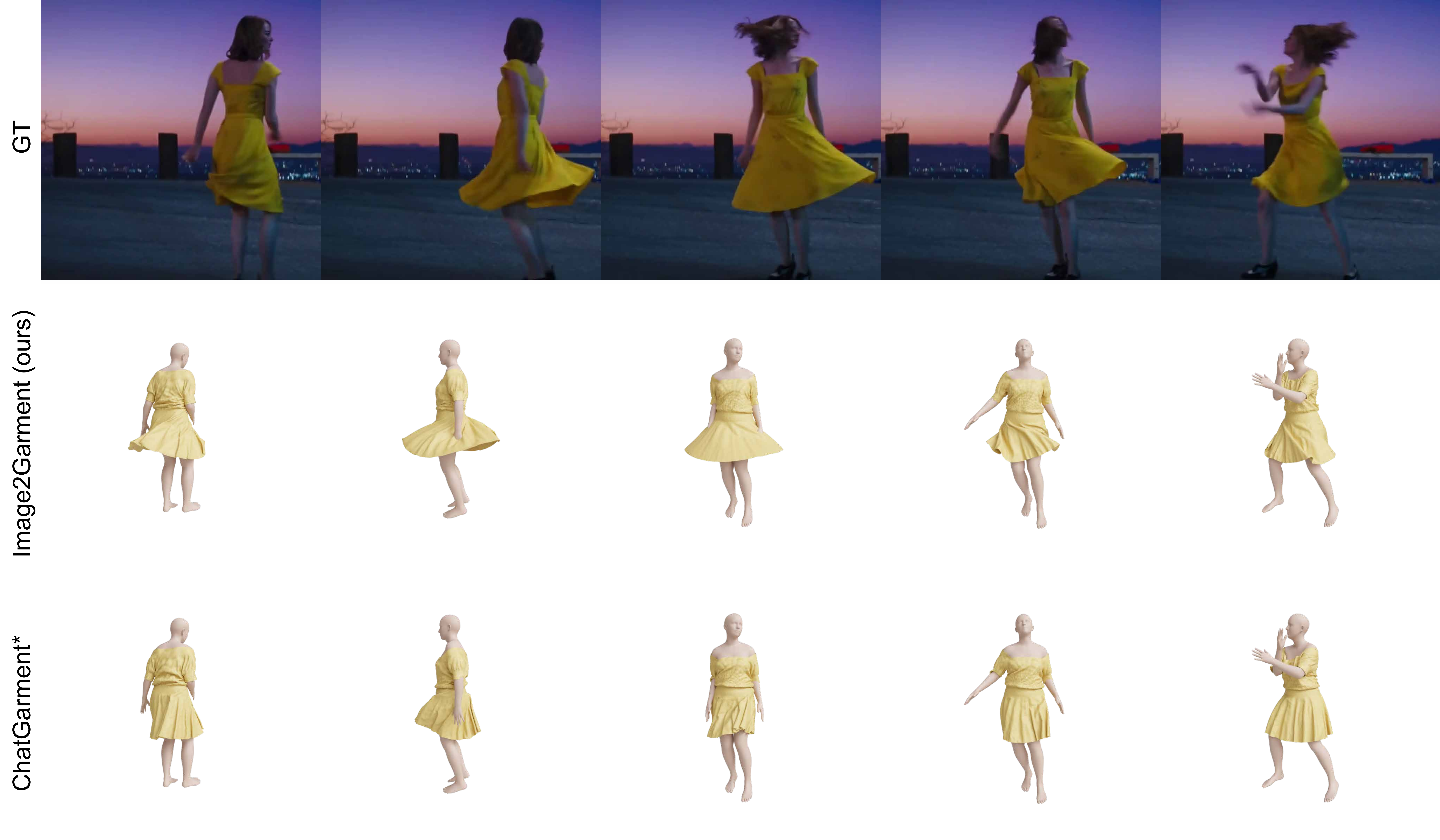}
    \caption{\textbf{Qualitative comparison on in-the-wild video.} The top row shows the original sequence. We only use the left-most frame as input. The rows below show renderings of the garments after simulation.}
    \label{fig:itw3}
\end{figure*}

\paragraph{Qualitative evaluation.}
In the supplementary figures, we provide additional in-the-wild examples produced by our Image2Garment pipeline. For each example, we show the original sequence (first image as input), the reconstructed sewing pattern, the final simulation frame, and close-ups of characteristic wrinkles. Figures~\ref{fig:itw1} and \ref{fig:itw3} compare our results against simulations generated using random fabric materials. Across diverse garment categories, our method produces more realistic silhouettes, time-consistent wrinkle patterns, and garment dynamics that closely match those observed in the source video clips. Please refer to our project website for the full video results.

\begin{figure*}
    \centering
    \includegraphics[width=0.8\textwidth]{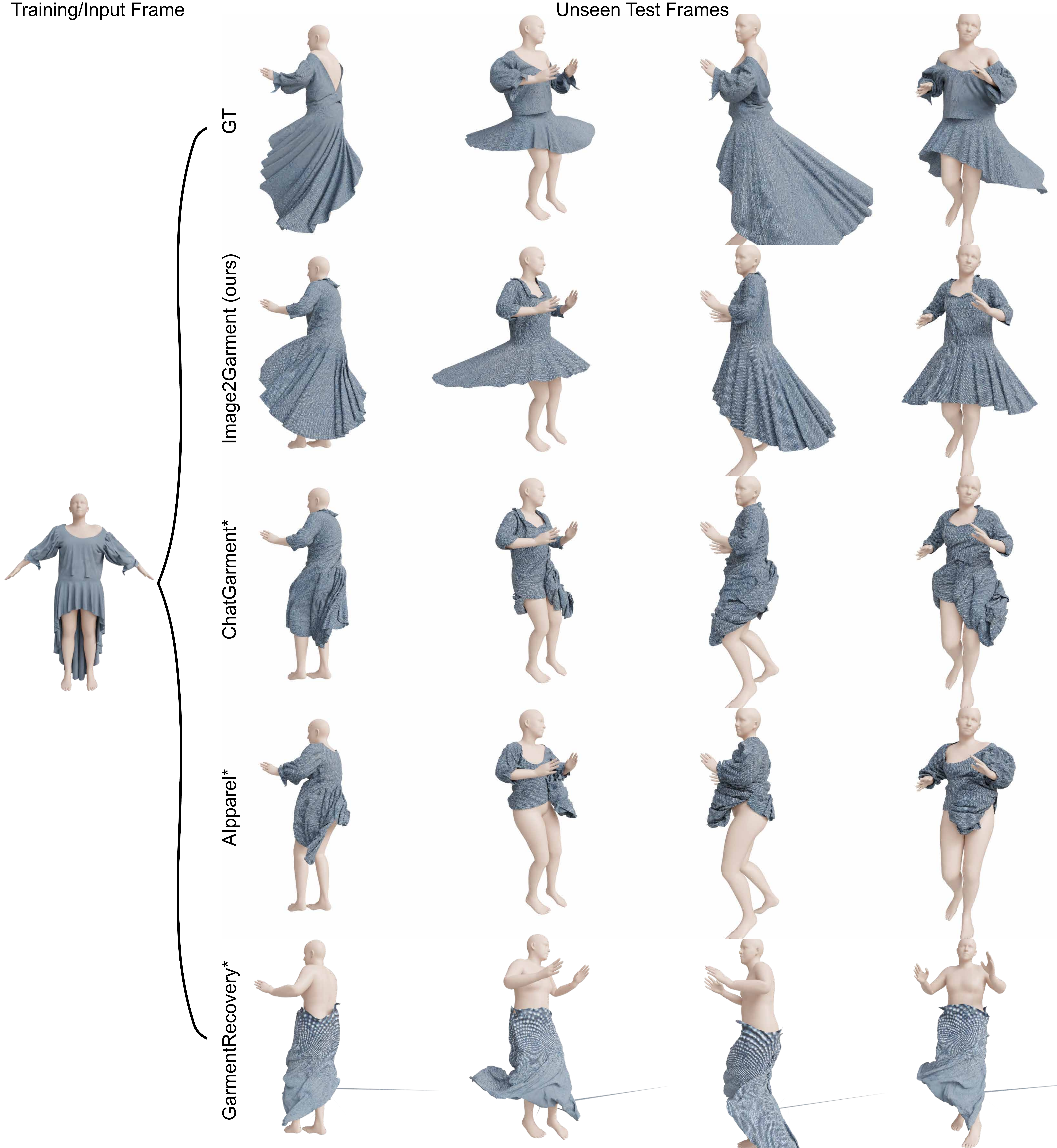}
    \caption{\textbf{Qualitative comparison on the northern spin example.}
    The left most column shows the input frames presented to each method and the columns on the right show the animated and rendered garments at different points in time of the animation. Our method's result most closely resemble the ground truth (top row).}
    \label{fig:sup_qual_2}
\end{figure*}

\begin{figure*}
    \centering
    \includegraphics[width=0.8\textwidth]{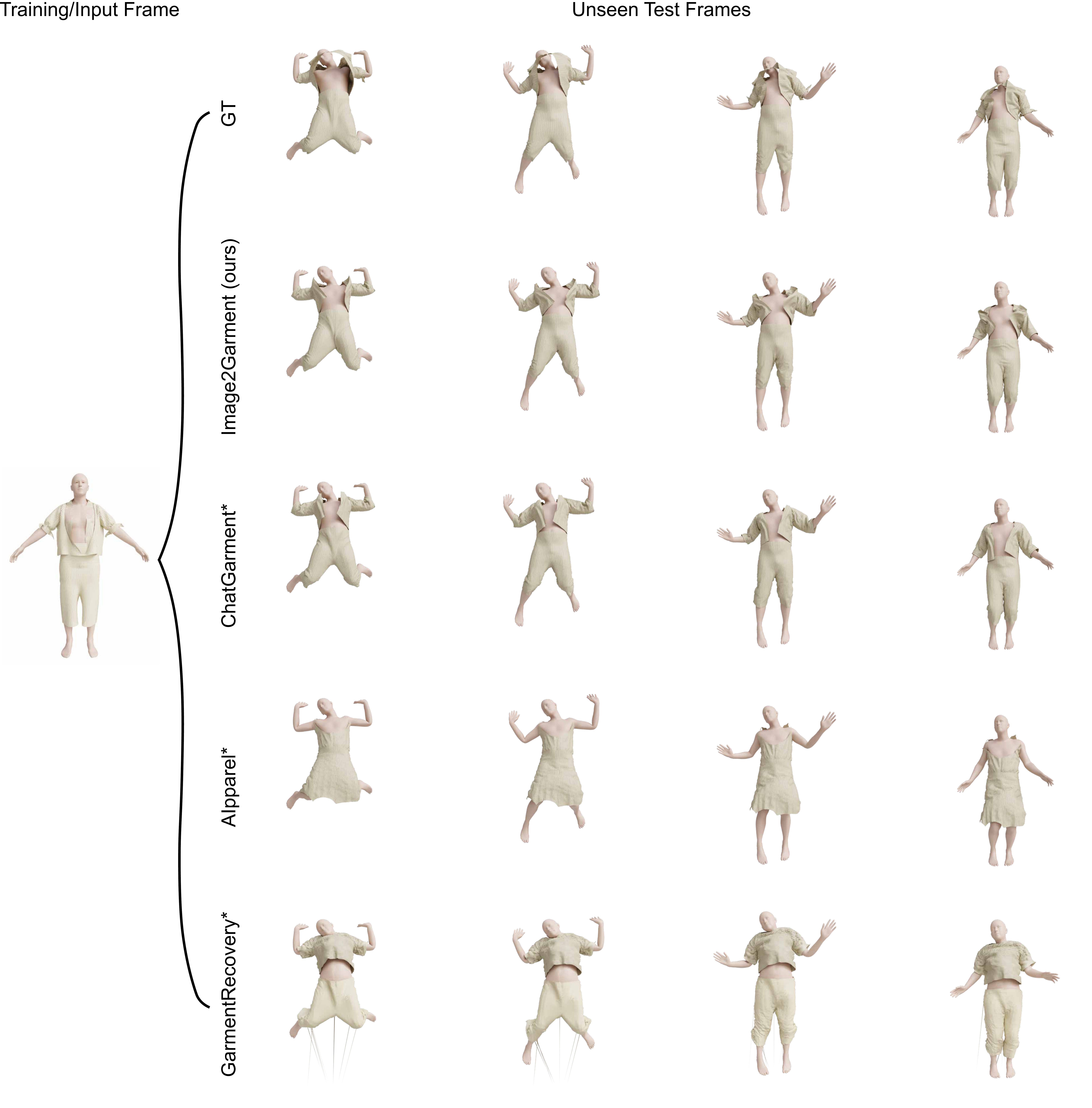}
    \caption{\textbf{Qualitative comparison on the joyful jump example.}
    The left most column shows the input frames presented to each method and the columns on the right show the animated and rendered garments at different points in time of the animation. Our method's result most closely resemble the ground truth (top row).}
    \label{fig:sup_qual_3}
\end{figure*}

\begin{figure*}
    \centering
    \includegraphics[width=0.8\textwidth]{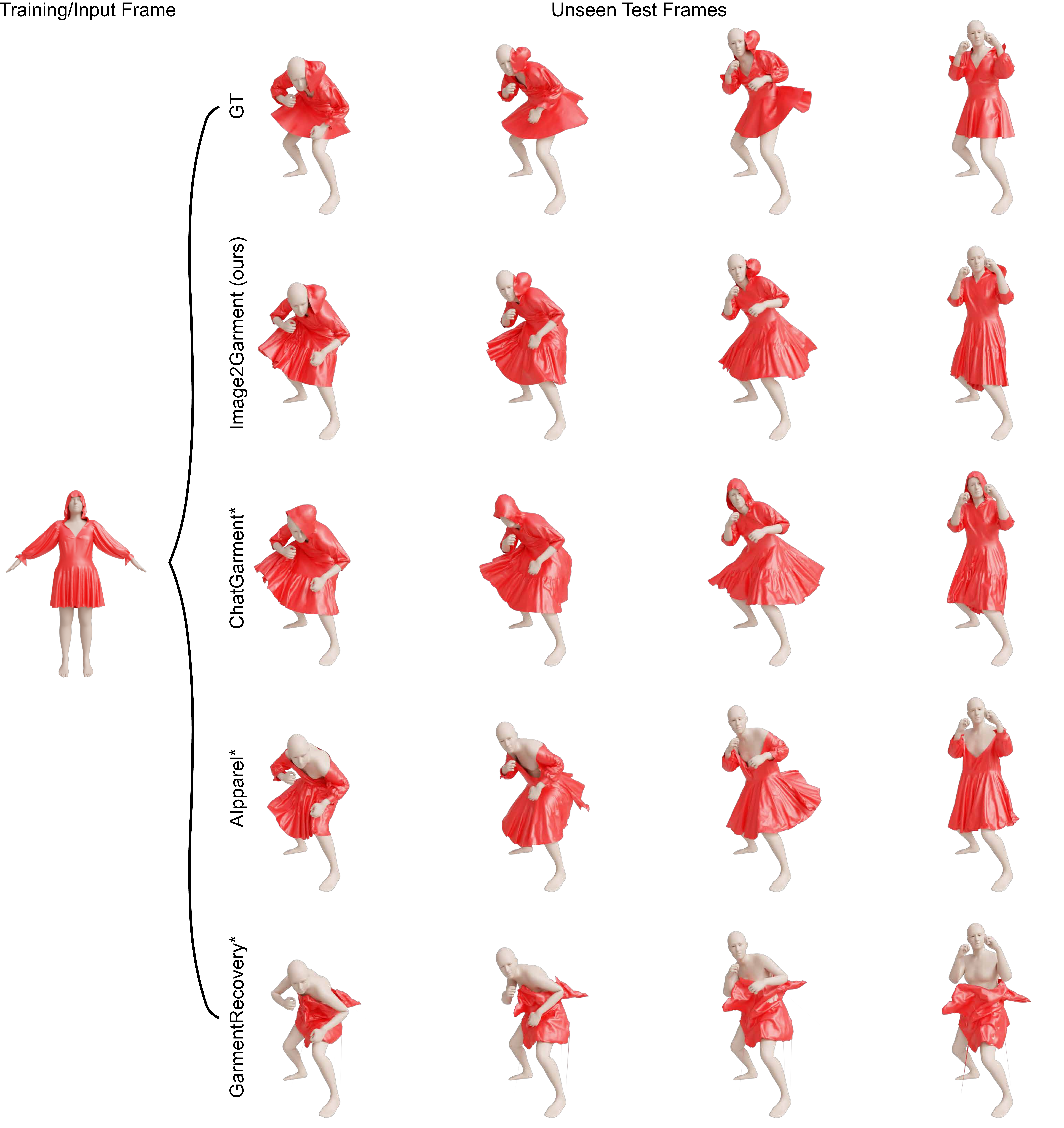}
    \caption{\textbf{Qualitative comparison on the hit reaction example.}
    The left most column shows the input frames presented to each method and the columns on the right show the animated and rendered garments at different points in time of the animation. Our method's result most closely resemble the ground truth (top row).}
    \label{fig:sup_qual_4}
\end{figure*}

Further, we present the remaining qualitative examples of the benchmark we created for evaluation. \cref{fig:sup_qual_2,fig:sup_qual_3,fig:sup_qual_4} show the input image in the left column and the draped and simulated garments on the right side. We can observe that our method produces the motions that most closely resemble the ground truth.


\subsection{Typical Failure Modes}

We observe several characteristic failure cases:
\begin{itemize}
    \item \textit{Ambiguous material appearance:} for visually ambiguous
    fabrics (e.g., cotton vs. viscose plain-weave), the model occasionally
    mispredicts the dominant fiber, leading to slightly over- or under-damped
    dynamics.
    \item \textit{Highly layered garments:} garments with multiple visible
    layers (e.g., jackets worn over dresses) are filtered out in FTAG, but
    in-the-wild images may still contain layering; in such cases, our single-layer
    physics may not fully capture the observed dynamics.
    \item \textit{Extreme accessories and trims:} heavy buttons, zippers, or
    appliqués are not explicitly modeled and can lead to local discrepancies
    between simulated and real wrinkles near those regions.
\end{itemize}

These cases suggest interesting future directions such as explicit layer
reasoning, accessory-aware physics modeling, and leveraging multi-view or
temporal cues when available.

\end{document}